\documentclass{article} 
\usepackage[final]{colm2026_conference}

\usepackage{microtype}
\usepackage{hyperref}
\usepackage{url}
\usepackage{booktabs}
\usepackage{xcolor}
\usepackage[utf8]{inputenc}
\usepackage{graphicx}
\usepackage{subcaption}
\usepackage{amsmath}
\usepackage{amsthm}
\theoremstyle{definition}

\usepackage{cleveref}
\usepackage{tcolorbox}  
\usepackage{multirow}
\usepackage{soul}
\usepackage{geometry}
\usepackage{booktabs}
\usepackage{amsthm}
\usepackage{float}
\usepackage{tabularx}
\tcbuselibrary{breakable}

\crefname{appendix}{Appendix}{Appendices}

\usepackage{lineno}

\definecolor{darkblue}{rgb}{0, 0, 0.5}
\hypersetup{colorlinks=true, citecolor=darkblue, linkcolor=darkblue, urlcolor=darkblue}

\title{Can LLMs Introspect? A Reality Check}

\author{Shashwat Singh, Tal Linzen, Shauli Ravfogel \\
Center for Data Science\\
New York University\\
\texttt{\{ss20428,linzen,shauli.ravfogel\}@nyu.edu} \\
}

\begin{document}

\ifcolmsubmission
\linenumbers
\fi

\maketitle

\begin{abstract}
Can large language models detect and report their own internal states? A number of recent studies have argued that they can. Drawing on lessons from human metacognition research, we argue that this conclusion may be premature. We identify two conditions that a paradigm needs to meet in order to establish introspection. First, the test needs to require \emph{privileged access}: it should not be solvable using cues available in the input. Second, it needs to require \emph{second-order computation}: second-order, meta-representations of first-order, task-related representations. This condition cannot be satisfied by task performance alone: it requires designs under which second-order and first-order accounts make divergent predictions. We re-examine two paradigms that have been used to argue for model introspection in light of these conditions. In the first, models must predict labels derived from their own hidden states; we find that classifiers that can only access the input match the models' in-context predictions, indicating that the original results do not demonstrate privileged access to internal representations. In the second paradigm, models must detect whether their internal states have been tampered with; we find they cannot reliably distinguish such interventions from manipulations of the input, suggesting that their success reflects generic anomaly detection rather than sensitivity to internal interventions in particular. We conclude that current evidence is insufficient to establish metacognitive monitoring in LLMs.

\vspace{0.5em}
\hspace{2.0em}\includegraphics[width=1.25em,height=1.15em]{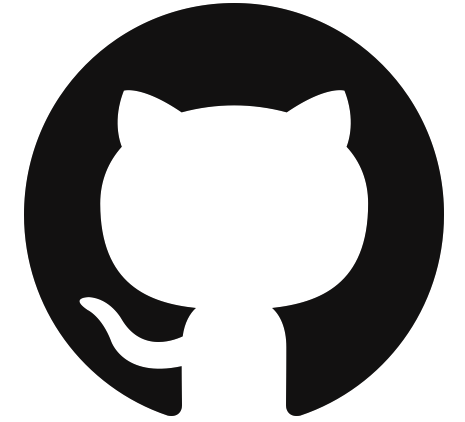}\hspace{.75em}
\parbox{\dimexpr\linewidth-7\fboxsep-7\fboxrule}{\url{https://github.com/shashwat1002/introspection_reality_check}}
\vspace{-.5em}

\end{abstract}

\section{Introduction}

Can language models reflect on their own internal processes? A surge of recent work asks whether LLMs can not only produce complex behavior but also monitor, report, and regulate their own internal states---abilities referred to in human cognitive science as introspection or metacognition \citep{nisbett1977telling, flavell1979metacognition, nelson1990metamemory}. Several recent studies have answered this question in the affirmative. Drawing on human metacognition research, which has yielded largely negative results and identified a range of confounds in self-report studies \citep{fleming2014measure}, we identify two necessary conditions for introspection in the strong sense we adopt: (i)~ \emph{privileged access}---the report must depend on internal states beyond what the input alone provides, or else self-report reduces to input-driven prediction of the model's own behavior \citep{shanahan2023role, turpin2023language}; and (ii) \emph{second-order computation}---the report must be produced by meta-representations that represent something \emph{about} first-order, task-related representations (e.g., ``my normal representation was manipulated,'' ``my answer was produced by guessing''). We argue that recent affirmative conclusions outrun the evidence on both counts: empirically, existing paradigms often fail to establish even the privileged access condition; and as a matter of construct validity, positive results in these paradigms underdetermine the second-order computation condition, since first-order and second-order processes can give rise to identical model self-reports.



We examine two prominent paradigms in light of these conditions. The first, which we term \emph{self-report classification}, seeks to establish introspection by showing that models can solve in-context learning tasks where the labels are derived from the models' own activations \citep{jian2025metacognitive, steinmetz2026belief}. For instance, models can predict labels corresponding to the first principal components of their representations \citep{jian2025metacognitive} or their tendency to follow parametric versus contextual knowledge \citep{steinmetz2026belief}. However, the fact that these labels are \emph{derived} by the experimenter from hidden states does not exclude the possibility that they can also be recovered by the model from the \emph{input alone}. Indeed, we show that labels in existing self-report classification tasks are often predictable directly from input features, without any access to the hidden states; and when the label are decorrelated from these features, performance collapses to near-chance. We conclude that self-report classification, as currently deployed, fails to satisfy the privileged access condition.

\begin{figure}[t]
    \centering
    \includegraphics[width=\linewidth]{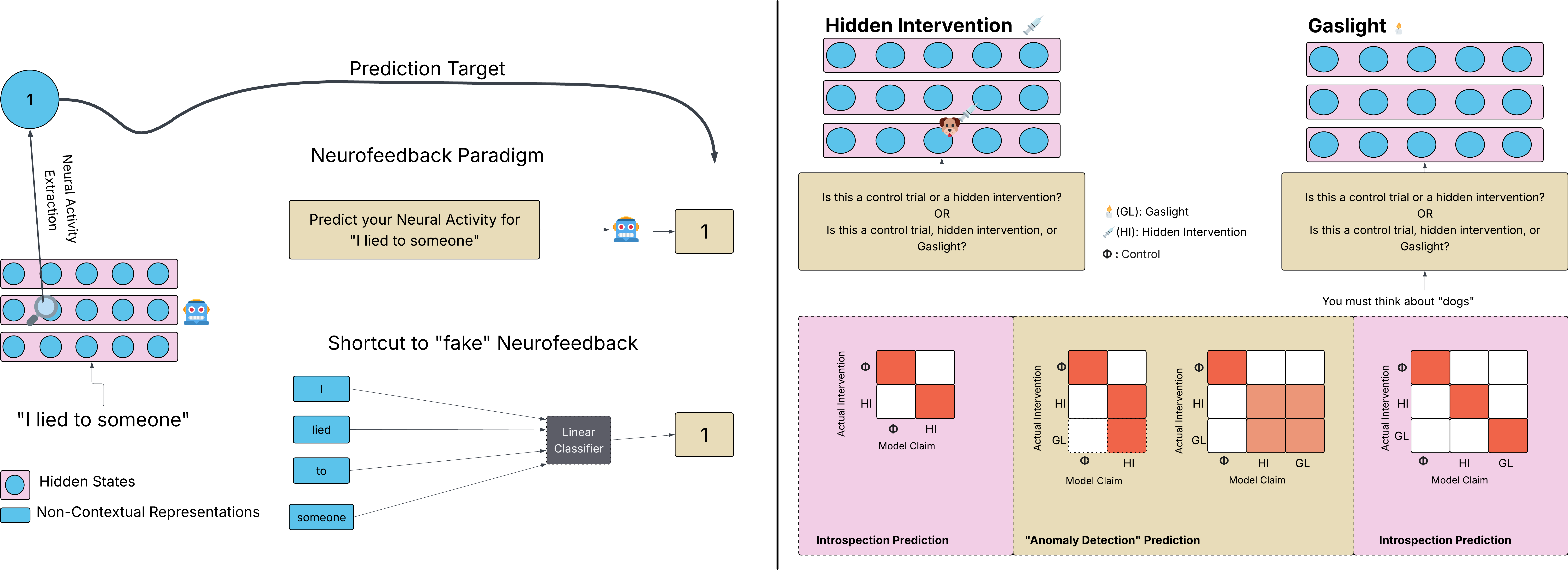}
    \caption{
\textbf{Input-controlled alternatives to purported introspection results.}
\textit{Left:} In the self-report paradigm of \citet{jian2025metacognitive}, labels are computed from a model's hidden state via a linear classifier or top PCA directions (A), then used as targets in in-context learning examples (B). Successful prediction has been interpreted as evidence of introspection. We show these labels are also predictable from uncontextualized input embeddings, so success need not imply privileged access. \textit{Right:} In the steering-awareness setting of \citet{lindsey2025emergent}, the anomaly detection hypothesis and introspection hypothesis make the same prediction and are confounded. Our design adds prompt interventions (the ``gaslight'' condition) matched to hidden-state interventions, separating the hypotheses: anomaly detection flags both as anomalous, while introspection selectively identifies hidden-state interventions.
}
    \label{fig:illustration_both}
\end{figure}

The second paradigm we study is \emph{intervention awareness} \citep{lindsey2025emergent}: in this study, Anthropic's Claude models were able to detect with nontrivial accuracy whether their activations were modified through steering, that is, by adding a vector representing a concept to the activations \citep{li2023inference, singh2024representation}. This paradigm satisfies the privileged access criterion by design. But, we argue, \citeauthor{lindsey2025emergent}'s conclusion that models can track interventions on the their internal states is too strong. In his design, which includes only normal and steered trials, ``my activations were intervened on'' and ``my current state is unusual'' are true on exactly the same trials, so accuracy alone cannot reveal which of the two explains the model's reports. We address this by adding a third ``gaslight'' condition that pushes the model toward a concept through the input prompt alone, with no direct intervention on the activations (\cref{expt1}). While in this condition the input might cause the model's state to be unusual, nothing is injected into its activations. We find that four open-weight models\footnote{We cannot replicate the paper directly, as the model tested by \citet{lindsey2025emergent} is not accessible outside of Anthropic.}  frequently incorrectly classify gaslight trials as injection trials, even when explicitly instructed to distinguish the two types of interventions (\cref{fig:illustration_both}, right). This suggests that the models' reports track that something is unusual rather than activation intervention specifically, and that there is no evidence for self-monitoring of the hidden states. We conclude that this paradigm leaves the second-order computation condition unaddressed; in fact, as we discuss in \cref{sec:construct-validity}, even success in our extended variant of the experiment would not meet this condition.

\label{sec:intro}


In summary, we conclude that current evidence is insufficient to establish that LLMs display strong metacognitive monitoring, and argue that future studies could be made more compelling by including stronger controls and, crucially, by pairing behavioral results with mechanistic evidence of a dissociable second-order process (for first steps in this direction, see \citealt{macar2026mechanisms}).

\section{Related Work}
\label{sec:related}

Work on LLM metacognition spans verbal calibration \citep{kadavath2022language, lin2022teaching, yona2024faithful}, probing for internal representations of confidence or truthfulness \citep{burns2023discovering, marks2024geometry, azaria2023internal, ravfogel2025emergence}, and paradigms testing whether models can report their own activation patterns \citep{jian2025metacognitive, steinmetz2026belief}; for an extended review, see \cref{app:related-approaches}. The human metacognition literature, which is rife with negative results, provides essential context for our work: people often confabulate explanations of their own behavior \citep{nisbett1977telling}, and apparent introspection into one's memorization abilities can stem from shallow cues such as familiarity rather than access to memory traces \citep{koriat1997monitoring}. Because above-chance correlations between confidence and accuracy can arise from first-order evidence alone, measures of metacognitive sensitivity have been developed within signal-detection theory \citep{maniscalco2012signal, fleming2014measure}, and first- and second-order accounts of confidence are differentiated by behavioral model comparison \citep{fleming2017self}. The parallel debate in comparative psychology is instructive: behavioral evidence of uncertainty monitoring in animals \citep{smith2003comparative, hampton2001rhesus} was met with first-order reinterpretations \citep{carruthers2008meta}, and progress ultimately required converging behavioral and neural evidence \citep{kepecs2008neural}.

Recent work has begun controlling for confounds in LLM self-report (\cref{app:related-controlled}). \citet{binder2024looking} show that a model predicts its own behavior better than an equally informed external model, but this capability is finetuned and, as they note, is consistent with the model \emph{simulating} its own forward pass---which satisfies the privileged access but not the second-order condition. \citet{plunkett2025self} finetune models to use randomly generated importance weights for different features in a classification problem, and find that they report those weights accurately, with strong controls against input-driven inference; the authors note that the reports may nonetheless reflect stored self-knowledge rather than real-time monitoring. \citet{song2025privileged} propose that to demonstrate introspection ones needs to show  a reliability advantage over any process of equal or lower computational cost available to a third party; they show that apparent successes can fail to meet this criterion. We share the motivation of the controls proposed in these studies, but we go further and argue that privileged access  on its own does not suffice to establish introspection in the strong sense.

Several groups have revisited the concept-injection detection work of \citep{lindsey2025emergent} in open-weight models, in some cases with explicit finetuning for the task \citep{vogel2025small, rivera2026steeringawarenessdetectingactivation}; we focus on pretrained models, since finetuning may equip the model with a task-specific mechanism rather than reveal a general capacity (\cref{app:related-scope}). Closest to our critique, \citet{lederman2026dissociating} argue that injection detection is content-agnostic: models detect that an anomaly occurred but cannot identify the injected concept. Their argument and ours target different links in the same chain: they dissociate detection from identification, whereas we ask what detection itself consists in---a second-order representation of the model's states, or a first-order sensitivity to irregularity (\cref{app:related-replications}).

\section{Construct Validity of Introspection Paradigms}
\label{sec:construct-validity}

\textbf{Defining introspection.} Introspection is not a univocal notion. On one family of views, a system knows its own states by using \emph{meta-representations}: states whose content attributes a property to a first-order representation, whether modeled on perception, as a kind of inner sense that scans first-order states \citep{armstrong1968, lycan1996}, or on thought, as a higher-order belief about them \citep{rosenthal2005consciousness}, possibly via a dedicated monitoring mechanism \citep{nichols2003mindreading}.\footnote{We treat the ability to dissociate the introspective mechanism from the first-order mechanism as the empirically tractable \emph{signature} of meta-representation, not as part of its definition, because some second-order accounts of introspection remain agnostic on dissociability.} On another, self-knowledge is obtained indirectly: through the same inferential processes used to attribute states to others \citep{carruthers2011opacity}, or through ``transparent'' procedures that answer ``do I believe $p$?'' by considering whether $p$ is true \citep{byrne2018transparency}. Our critique targets the first family: the claim that LLMs possess a dedicated capacity to inspect their own hidden states, over and above ordinary forward-pass computation. The inferential view of introspection is easier to satisfy and is not what motivates claims of emergent introspective awareness. We do not attribute the strong notion to the authors we examine, who define their targets functionally; our claim is that functional success does not license the stronger reading that terms such as ``introspective awareness'' invite.

\textbf{Privileged access is necessary but not sufficient.} Any paradigm advanced as evidence of introspection must satisfy the privileged-access condition \citep{song2025privileged}: the labels must depend on features of the hidden states that are not recoverable from the input alone. But privileged access is not sufficient. Every computation in a language model is performed over hidden states, so a task whose labels depend on $h(t)$ need not engage machinery distinct from ordinary forward-pass computation. Sentiment analysis is a readout of hidden states, yet no one takes it to show that the model introspects on its sentiment representations; a label defined over $h(t)$ is, by itself, no different. A state that merely covaries with a first-order representation is not thereby \emph{about} it.

What would count, then? Consider \emph{reflection} in programming languages. An ordinary program uses values in memory; a reflective program can also obtain descriptions of those values and act on them. Transformers always use their hidden states, but the second-order computation condition asks whether they represent those states in this second sense. We cannot settle this question simply by measuring the accuracy of the system's reports. A rule that prints ``my activations were altered'' whenever a state norm exceeds a threshold could reliably detect interventions through a direct causal pathway, without monitoring the state as such, producing identical behavior to a genuine self-monitoring mechanism. Distinguishing these two mechanisms requires not only designs on which they make different predictions of the kind developed in human metacognition research \citep{fleming2014measure, fleming2017self}, but also, ultimately, evidence about the model's internal organization; for a more detailed discussion, see \cref{app:meta-representation}.

\section{Experiments}
\label{sec:experiments}

Based on the discussion above, we reassess whether results from the self-report classification and intervention awareness paradigms support the claim that models can introspect.

\subsection{Self-report classification: the paradigm of \citet{jian2025metacognitive}}
\label{expt-biofeedback}
 
\subsubsection{Method}
 
The self-report paradigm (originally referred to as ``biofeedback'', in analogy to the psychological technique) derives a proxy from the language model's activations and uses it as the prediction target in an in-context learning task presented in a fresh run, with new inputs.\footnote{The term is borrowed from a technique in which humans are shown real-time readouts of their own physiological signals and learn to regulate them through feedback.} \citet{jian2025metacognitive} derive such proxies in two ways, which we refer to as \emph{supervised} and \emph{unsupervised}. The \textbf{supervised} proxy is derived from a logistic regression probe trained on the model's hidden states as it processes a labeled dataset; the inner products between the probe and the hidden states are clustered into two bins, which constitute the classification labels. The \textbf{unsupervised} proxy is based on principal component analysis (PCA): the inner products between each sample's representation and a principal component are clustered into two bins, and the resulting assignments to bins serve as prediction targets. For both proxies, the model is given a set of in-context examples and evaluated on held-out samples. Following \citeauthor{jian2025metacognitive}, we use the commonsense subset of the Ethics dataset \citep{hendrycks2023aligningaisharedhuman}, which consists of first-person descriptions of actions labeled for whether they constitute an ethical violation; for instance, ``I told my baby I loved her when she cried'' does not contain a violation, whereas ``I told my baby I hated her when she cried'' does.
 
\citeauthor{jian2025metacognitive} take successful prediction of these proxy labels as evidence of introspection. We ask whether above-chance performance could instead arise from the model picking up on an underlying classification task implied by the labels: in the supervised setting, this could be the original task of detecting ethical violations; in the unsupervised setting, any other semantically meaningful property of the materials---or even a superficial one such as sentence length---that correlates with the labels defined by the principal component. We address the two settings separately, evaluating \texttt{Llama-3.1-8B-Instruct} (following \citeauthor{jian2025metacognitive}) and, additionally, \texttt{Llama-3.1-70B-Instruct}.
 
\paragraph{Removing semantic correlates in the supervised setting.} We retain the original data but randomly permute the labels before the probe is trained. This breaks the correspondence between input semantics and probe output while preserving the label distribution. The probe trained on permuted labels still defines a well-formed---if arbitrary---direction in hidden-state space, one that could in principle be predicted perfectly, and therefore remains a valid proxy of the model's neural activity for testing privileged access. This control is related to \citeauthor{jian2025metacognitive}'s own analysis of higher-order principal components, which they probe on the grounds that these lack clear semantic content; random relabeling instantiates the same intuition without relying on the assumption that higher-order components do not have a semantic interpretation.
 
\paragraph{Uncovering input-based shortcuts in the unsupervised setting.} For the PCA variant we use a different control: we fit linear probes on the mean-pooled layer-0 representations of the inputs to predict the binary-clustered PCA components of a given hidden layer, repeating this for each layer and reporting the average accuracy across layers. Successful layer-0 probes would indicate that the labels are predictable directly from input features, such that high accuracy on the in-context task can be achieved without introspective access. For full experimental details, see \cref{app:expt2_dirty_details}.

\begin{figure}[t!]
    \centering
    \begin{subfigure}[t]{0.40\textwidth}
        \centering
        \includegraphics[width=\textwidth]{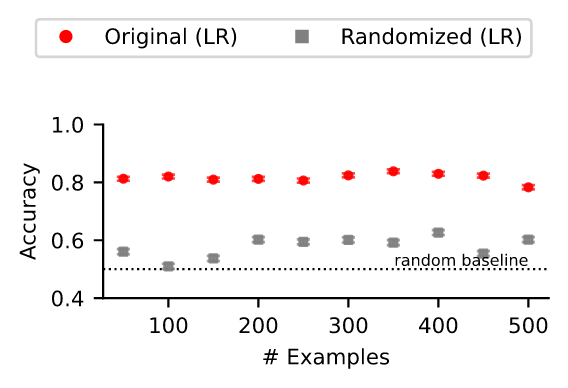}
        \caption{\texttt{Llama-3.1-8B}: logistic regression probe.}
        \label{fig:relabel_probe_8}
    \end{subfigure}
    \hfill
    \begin{subfigure}[t]{0.54\textwidth}
        \centering
        \includegraphics[width=\textwidth]{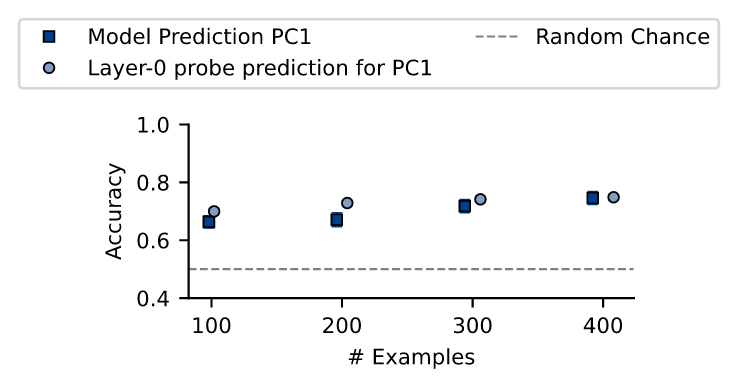}
        \caption{\texttt{Llama-3.1-8B}: PCA.}
        \label{fig:pca_probe_8}
    \end{subfigure}

    \caption{Controls for the \citet{jian2025metacognitive} paradigm. The $x$-axis is the number of in-context examples for the original setting, which equals the number of training samples for our probes. \emph{Left:} accuracy drops from the original labels (red) to near the majority-class baseline after random relabeling (grey), which removes semantic correlations that could be inferred from the input. \emph{Right:} probes trained on layer-0 inputs alone (circles) match or exceed the models' own in-context predictions of the PCA-derived labels.}
    \label{fig:biofeedback_control}
\end{figure}
 
\subsubsection{Results}
 
\paragraph{Models struggle to predict proxies decorrelated from semantics.} While accuracy on the original, semantically aligned labels is considerably above chance, accuracy on the arbitrary direction defined by a probe trained on randomly relabeled data drops to close to the majority-class baseline (\cref{fig:biofeedback_control}a,b). Performance in the original paradigm therefore need not require access to the model's hidden states, and is consistent with in-context learning of the semantic regularities present in the data. Experiments with \texttt{Llama-3.1-70B} and \texttt{Qwen-2.5-7B-1M} yield similar results (see Appendix~\ref{app:biofeedback}).
 
\paragraph{PCA-derived labels are linearly predictable from input features.} We report the analysis for the first principal component; results for other components are similar. The layer-0 probes closely track the models' in-context performance across training sizes (\cref{fig:biofeedback_control}), indicating that the task can be solved from input-embedding features alone, without privileged access to any layer's activations. We report similar results for \texttt{Llama-3.1-70B} in Appendix~\ref{app:biofeedback}.
 
\paragraph{Semantic correlates present a confound in the neural-control paradigm as well.} In another experiment, \citeauthor{jian2025metacognitive} instruct models to \emph{control} the value of the neural proxy presented in context; they observe that the models are able to so to some extent. But since their prediction experiments do not rule out the possibility that the models are using semantic properties of the input to solve the task, these results are equally consistent with being able to control generation, a capability we know models have.
 
\subsection{Self-report classification: the Belief Dominance framework of \citet{steinmetz2026belief}}
\label{expt3}
 
\subsubsection{Method}
 
\citet{steinmetz2026belief} study introspection by assessing how models resolve conflicts between competing ``beliefs'': the model's parametric factual knowledge (the \textsc{base} entity), which is known to be correct, and contradicting evidence presented in context (\textsc{counter}). They use an augmented version of CounterFact \citep{meng2022locating}, which pairs prompts with factual and plausible counterfactual completions---e.g., ``The capital of France is'' would have Paris as the \textsc{base} and Rome as the \textsc{counter}---and supply the conflicting evidence in several forms (see \cref{app:table_augmentations} in the Appendix). Their Belief Dominance (BD) metric measures the relative strength of the two beliefs across layers (for details, see \cref{app:bd}); BD values are discretized into three clusters by $K$-means, and the model is asked to predict these cluster labels from in-context examples. Successful prediction is taken to indicate that the model can introspect on its beliefs.
 
We ask whether the cluster labels are predictable from input features alone. We train linear probes on the concatenated layer-$0$ embeddings of two entities, in two settings: the subject and the \textsc{counter} (``France'' and ``Rome''), or the \textsc{base} and the \textsc{counter} (``Paris'' and ``Rome''). Crucially, the probes are not shown any other part of the prompt and receive no indication that it contains a factual error that could lead to a belief conflict. We split \citeauthor{steinmetz2026belief}'s $900$ examples evenly into train and test sets across $15$ random splits, balancing the training set across the three classes, and report mean accuracy and standard deviation (see \cref{app:data_distribution_expt_3} for details). \citeauthor{steinmetz2026belief} use a balanced set of $30$ in-context samples, and we report their numbers as published. Since the training set is explicitly balanced in all three settings---the original in-context setting and our two layer-0 probe. Following \citeauthor{steinmetz2026belief}, we experiment with \texttt{Llama-3-70B} and \texttt{Gemma-3-27B}.

\begin{table}[t]
\centering
\small
\setlength{\tabcolsep}{4pt}
\renewcommand{\arraystretch}{1.05}
\begin{tabular}{p{4.2cm}cccc}
\toprule
\textbf{Setting} & \multicolumn{2}{c}{\textbf{\textsf{Llama 3-70B}}} & \multicolumn{2}{c}{\textbf{\textsf{Gemma 3-27B}}} \\
 & \textbf{BD(base)} & \textbf{BD(counter)} & \textbf{BD(base)} & \textbf{BD(counter)} \\
\midrule
ICL \citep{steinmetz2026belief}
& 0.46 {\scriptsize$\pm$ 0.02}
& 0.54 {\scriptsize$\pm$ 0.05}
& 0.48 {\scriptsize$\pm$ 0.02}
& 0.42 {\scriptsize$\pm$ 0.04} \\
Probe (subject + counter)
& \textbf{0.50 {\scriptsize$\pm$ 0.03}}
& 0.55 {\scriptsize$\pm$ 0.03}
& \textbf{0.57 {\scriptsize$\pm$ 0.02}}
& \textbf{0.50 {\scriptsize$\pm$ 0.02}} \\
Probe (base + counter)
& 0.49 {\scriptsize$\pm$ 0.02}
& \textbf{0.57 {\scriptsize$\pm$ 0.02}}
& 0.55 {\scriptsize$\pm$ 0.02}
& 0.48 {\scriptsize$\pm$ 0.01} \\
\bottomrule
\end{tabular}
\caption{Accuracy on Belief Dominance cluster labels. Layer-0 entity probes match or exceed the in-context performance reported by \citet{steinmetz2026belief}. Training sets are balanced across the three classes in all settings (the test set is in the natural distribution in this figure); results on a balanced test set are in \cref{tab:bd_results_balanced}.}
\label{tab:bd_results}
\end{table}
 
\subsubsection{Results}
 
\paragraph{Belief Dominance labels are linearly predictable from input features.} We find that linear probes that only have access to the entities' uncontextualized embeddings match or surpass the models' in-context performance on the BD prediction task (\cref{tab:bd_results}). The information used to solve this task is thus largely predictable from properties of the entities alone, without access to the model's internal representations or to any contextual information about the belief conflict. We hypothesize that this reflects simple properties such as entity frequency, in line with the documented relationship between frequency and both factual recall and context-sensitivity \citep{kandpal2023large, mallen2023trust, xie2024adaptive}.
 
\paragraph{Input-driven computation can also explain the results of intervention experiments.} In a separate experiment, \citeauthor{steinmetz2026belief} report that injecting a vector corresponding to the \textsc{counter} entity increases the BD(\textsc{counter}) and lowers thed BD(\textsc{base}) predicted by the model, and take this as evidence of metacognitive monitoring. But since both quantities are predictable from the layer-0 embeddings of the entities, this finding is equally consistent with the possibility that the model is computing a function of those entities that is the perturbed by the intervention. The fact that intervening on a representation changes behavior establishes that the representation is causally efficacious, in line with the broader literature on the causal encoding of concepts in language model representations, but does not show not that the model has introspective access to the representation.
 
\subsection{Intervention awareness: can models detect activation steering?}
\label{expt1}
 
\subsubsection{Method}
 
In the ``thought injection'' paradigm of \citet{lindsey2025emergent}, the prompt tells the model that a researcher can inject ``thoughts'' into its ``mind'' and that some trials will contain an intervention while others are controls; a low rate of false positives (cases where the model reports an intervention when there was none), together with a non-trivial true positive rate, is then taken as evidence for introspection. Crucially, the model is not asked to discriminate between types of perturbation, but only to distinguish a normal forward pass from one in which an internal perturbation occurred (a \emph{two-way} design). This leaves open the possibility that the model succeeds by detecting that the trial is irregular or out-of-distribution in a general way---a capacity that does not specifically indicate that the model is capable of second-order processing of its own hidden states.\footnote{We do \emph{not} identify the existence of second-order representation with source monitoring, the ability to detect whether a signal arose from the input or from a hidden-state intervention: a genuine monitor could represent the content of its states while being oblivious to their provenance, and, conversely, source-specific reports are achievable through first-order routes such as signature detection. We claim that whether or not any introspective monitoring is involved, there is no evidence that the reports track interventions on the model's hidden states, which is what the term ``steering awareness'' implies.}
 
\textbf{Three-way design.} To address this issue, we introduce a third condition, an \emph{input-level intervention} we call the ``gaslight'' condition, in which the prompt instructs the model to steer the conversation toward a concept---for example, \emph{You are obsessed with X. Everything you want to say somehow relates to X}. The gaslight string is put in the user string right before the prompt that describes the experimental setting (see Appendix \ref{app:all_prompts} for experimental details and full prompts). Crucially, this condition involves no intervention on the model's activations. It allows us to ask two questions. First, if we restrict the model to the two responses that were available to it in \citeauthor{lindsey2025emergent}'s study---thought injection or no perturbation---how does the model classify input-level interventions? Second, if we give it three options---textual manipulation, intervention on the model's ``thoughts'', and no intervention---can it distinguish input-level from activation-level interventions? Prompts appear in \cref{app:prompt1_2set} and \cref{app:prompt1_3set}.
 
\textbf{Models.} We experiment with the following models: \texttt{Llama-3.1-70B}, \texttt{Llama-3.1-8B} \citep{grattafiori2024llama3herdmodels}, \texttt{Qwen-2.5-72B} \citep{qwen2025qwen25technicalreport}, \texttt{Qwen-3-32B} \citep{yang2025qwen3technicalreport}, and \texttt{Gemma-27b-it} \citep{gemmateam2025gemma3technicalreport}. We randomize the order in which response options are listed.
 
\textbf{Steering.} Interventions are implemented as linear steering: a direction encoding a target concept is added to the residual stream at inference time \citep{li2023inference, singh2024representation}. We apply the intervention at all of the positions of the string ``Trial 1: What do you detect?'', which ends the prompt, and search over layers and intervention strengths $\alpha$, reporting the setting at which each model detects injection most accurately (see \cref{app:search_space,app:sec_optimal_expt_1}). Except for \texttt{Gemma-3-27b-it}, we do not steer the sampled tokens (which \cite{lindsey2025emergent} do), since doing so renders the output unintelligible; this still reproduces \citet{lindsey2025emergent}, as the detection answer is the first sampled token. For \texttt{Gemma-3-27b-it} we follow the original protocol, since we find that steering the prompt alone has no causal effect on neutral prompts like ``tell me something''.
 
\begin{figure}[t!]
     \centering
     \begin{subfigure}[b]{0.42\textwidth}
         \centering
         \includegraphics[width=\textwidth]{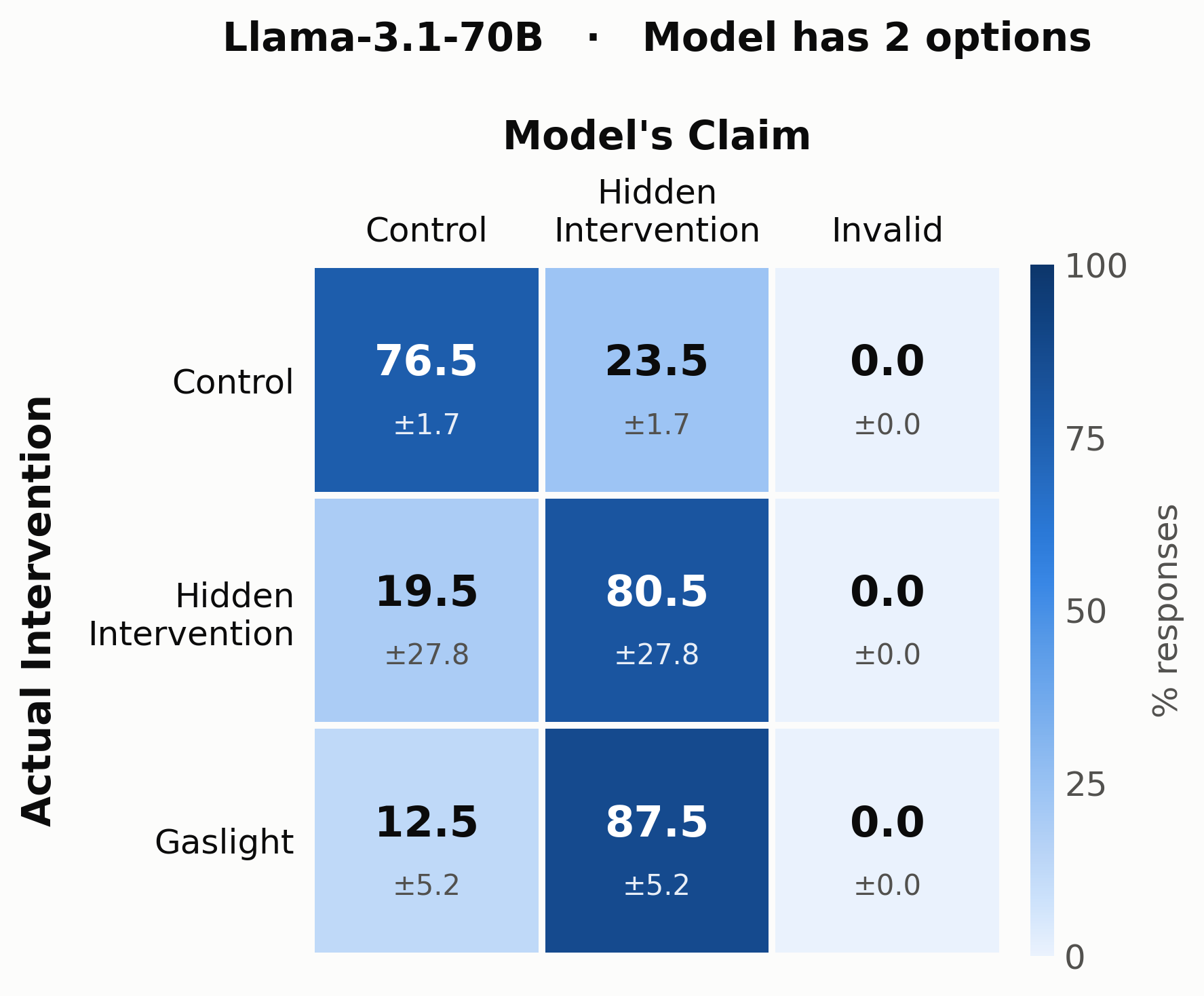}
         \includegraphics[width=\textwidth]{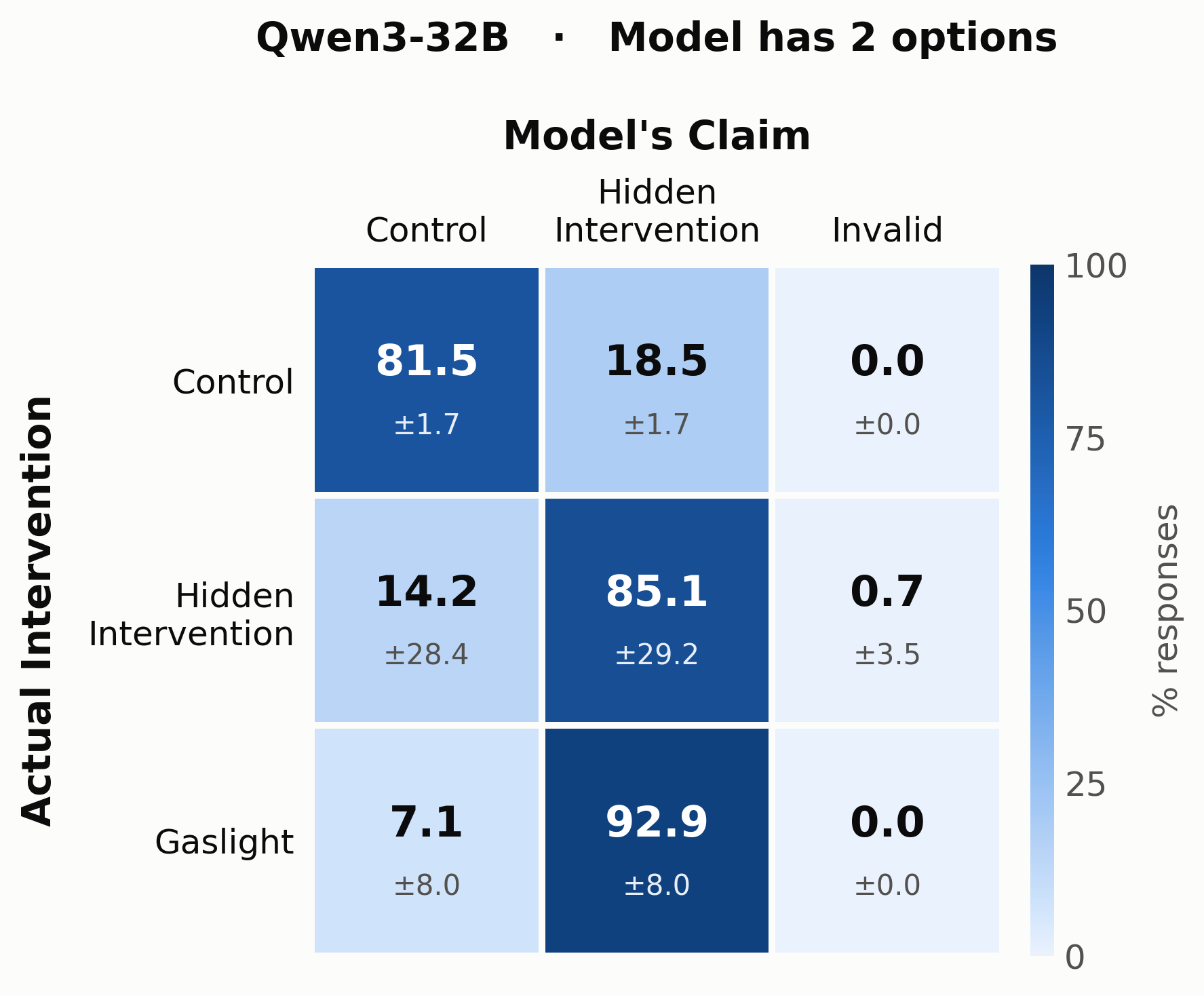}
         \caption{Two-way condition: the model is asked to choose between no intervention and activation intervention.}
         \label{fig:steering_2set}
     \end{subfigure}
     \hfill
     \begin{subfigure}[b]{0.5\textwidth}
         \centering
         \includegraphics[width=\textwidth]{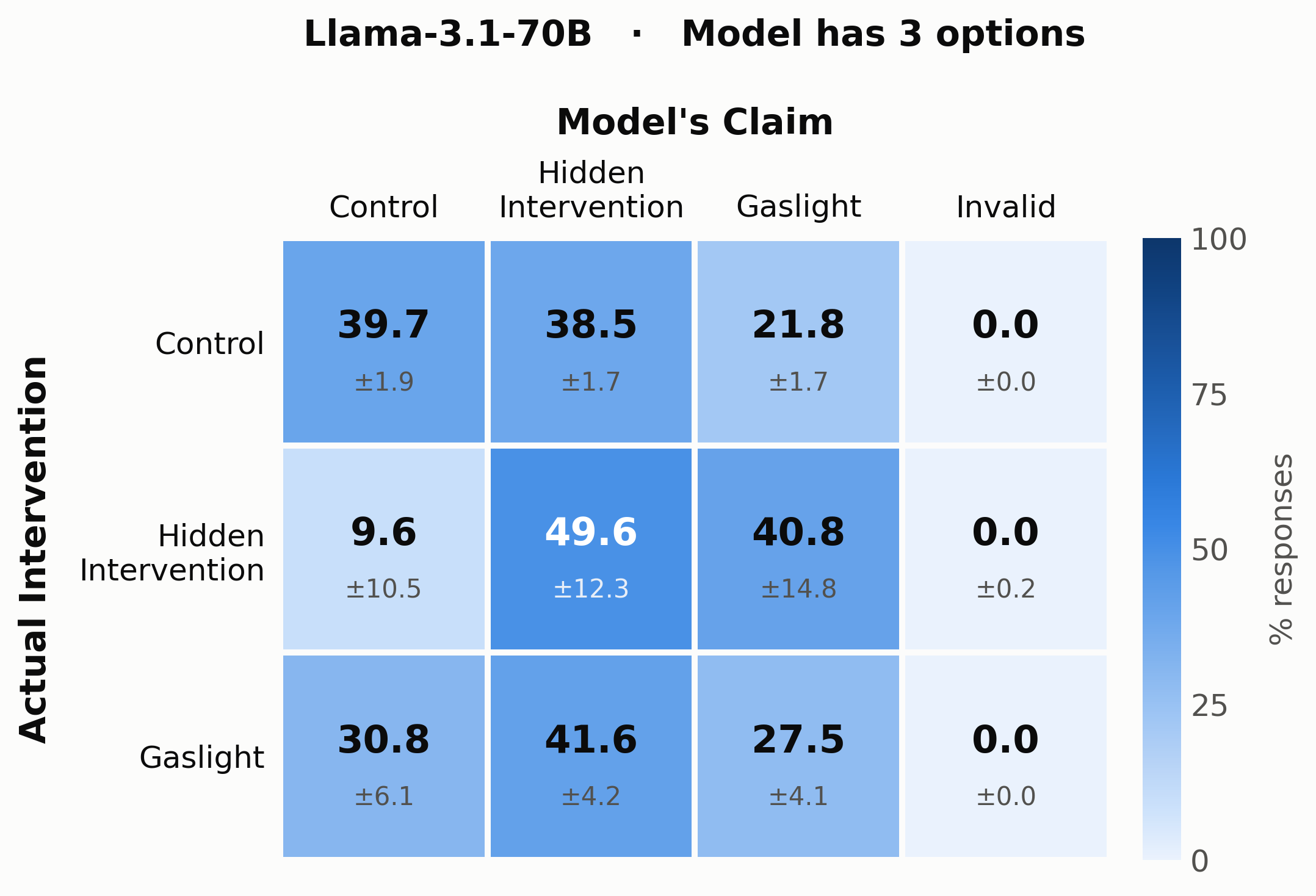}
         \includegraphics[width=\textwidth]{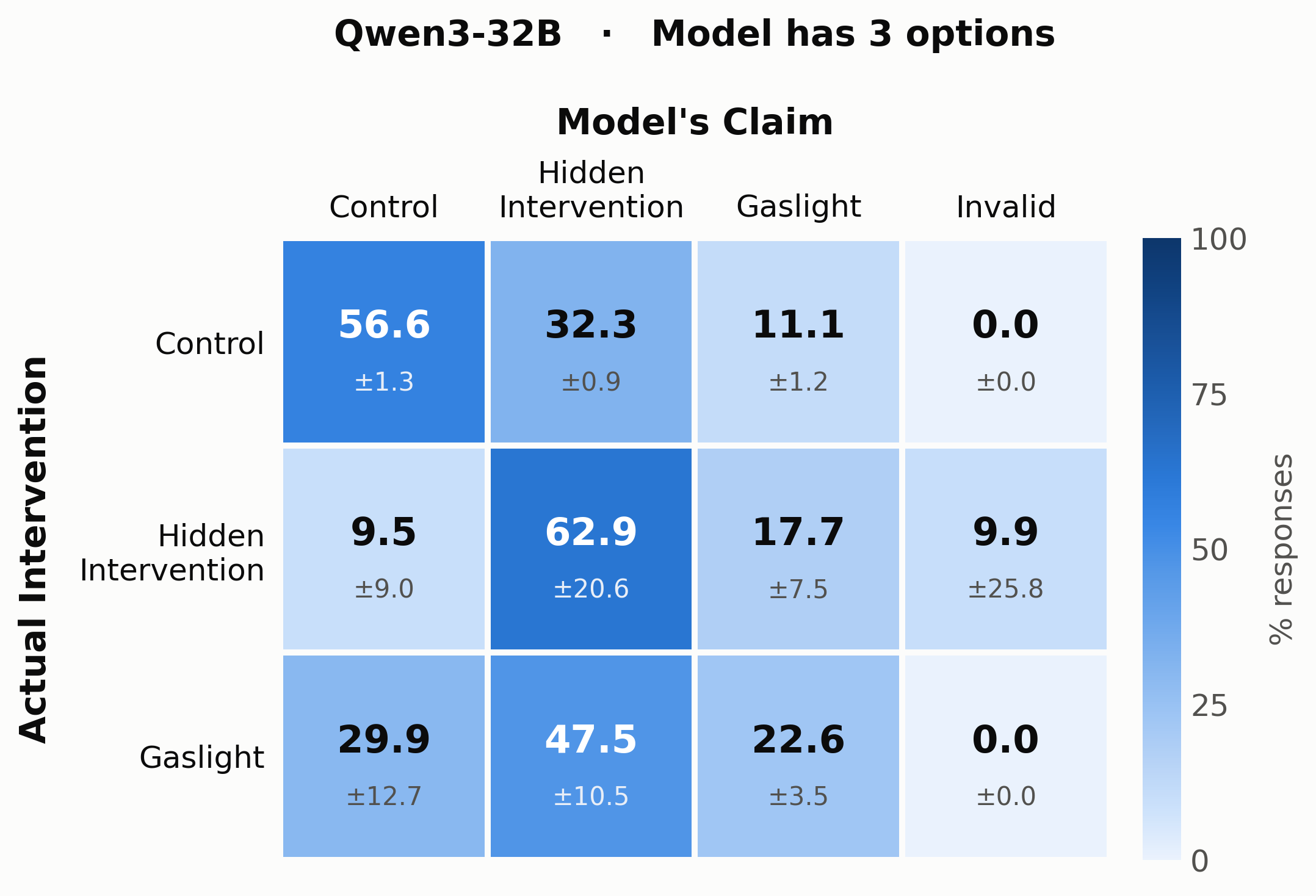}
         \caption{Three-way condition: the model chooses between no intervention, input intervention and activation intervention.}
         \label{fig:steering_3set}
     \end{subfigure}

    \caption{Response distributions for our extension of the \citet{lindsey2025emergent} intervention detection task. \emph{Left:}~In the two-way setting, \texttt{Llama-3.1-70B} and \texttt{Qwen-3-32B} achieve above-chance discrimination between control and activation-level intervention trials. \emph{Right:}~In the three-way setting, the model fails to reliably distinguish input-level from activation-level interventions, suggesting that the models' detection accuracy is compatible with a generalized sensitivity to  irregularity (if at all; in some cases, it is just showing a preference for a particular option regardless of the intervention type), and does not necessarily reflect introspective access to hidden states. The rows are normalized to 100\% and the standard deviations are calculated across the concepts listed in Appendix~\ref{app:concepts}.}
     \label{fig:steering_comparison}
\end{figure}

The gaslight condition is evaluated with 500 samples per concept, whereas the steering condition is evaluated with 500 samples in total across concepts for each combination of layer and intervention strength, due to computational constraints; the best-performing hyperparameter values are then rerun with 250 samples per concept. Detection rates vary considerably across concepts, which contributes to the large standard deviations for hidden interventions (for the per-concept breakdowns, see \cref{app:concept_breakdown}). Models are also sensitive to task wording, so we run two prompt variants (conversational and instruction-based) per setting and report, in \cref{fig:steering_comparison}, the variant that reproduces \citet{lindsey2025emergent} on the two-way task; all remaining prompts and results are in \cref{app:prompt_variation,app:all_prompts}. These are all consistent with the main results, i.e for all prompt wordings where the models succeed the two-way setting well enough to pass the criterion for reproducing \cite{lindsey2025emergent}, they fail the three-way setting. Unsurprisingly, models fail the three-way setting also in cases where they fail the two-way setting.

\subsubsection{Results}
 
In the two-way setting, \texttt{Llama-3.1-70B} and \texttt{Qwen-3-32B} (\cref{fig:steering_comparison}, left) reproduce the pattern reported by \citet{lindsey2025emergent} for Claude: very few false positives on control trials and a non-trivial detection rate on hidden-intervention trials. In fact, if we take these results to constitute evidence of introspection, this would indicate that \texttt{Llama-3.1-70B-Instruct} has substantially stronger introspection capabilities than the version of Claude evaluated by \citeauthor{lindsey2025emergent}. By contrast, \texttt{Qwen2.5-72B} almost never reports an intervention in any condition and does not reproduce the effect (see \cref{app:qwen_25_72}).
 
Critically, \texttt{Llama-3.1-70B} and \texttt{Qwen-3-32B} also label gaslight trials as activation interventions (third rows of the heatmaps in \cref{fig:steering_comparison}, left; recall that this setting does not provide the model with the option of reporting an input intervention). Their responses on activation-intervention trials are therefore equally compatible with a general capacity to detect an anomalous input. The three-way setting bears this out: given an explicit input intervention response option, \texttt{Llama-3.1-70B} performs near chance, with probability concentrated disproportionately on the activation-intervention option. \texttt{Qwen-3-32B} likewise fails to separate the two intervention types, showing a strong preference for the control response. \texttt{Gemma-3-27B} reproduces \citeauthor{lindsey2025emergent}'s results to a large extent in the two-way setting, but fails in the three-way setting (see Appendix~\ref{app:gemma_expt1}).
 
\textbf{Discussion.} \citeauthor{lindsey2025emergent}'s original experiment shows that some models correctly report steering interventions, but does not rule out the possibility that these reports rely on a general sensitivity to irregularity. Our design tests this directly. While two of the three models we tested replicate \citeauthor{lindsey2025emergent}'s results, these models frequently classify gaslight trials as injections, and, given the option to report a prompt manipulation, still fail to separate the two intervention types. Both observations are consistent with sensitivity to generic irregularity and require no introspective access to hidden states. We therefore conclude not that these models demonstrably lack introspective capacities---models could likely be trained to separate the two interventions---but that success in the two-way paradigm cannot, on its own, be taken as evidence of such capacities.

\section{Conclusions}
\label{sec:conclusions}

We have examined two paradigms taken to provide evidence for metacognitive monitoring in LLMs, and found both wanting---but in different ways. Self-report classification tasks fail to establish even privileged access: their labels are largely predictable from the input alone, and performance collapses to baseline once the labels are decorrelated from input semantics. The intervention awareness paradigm satisfies the privileged access condition by construction, yet models frequently report prompt-level manipulations as injected thoughts and, given the option to distinguish the two types of intervention, fail to do so---a pattern fully accounted for by a generic sensitivity to irregularity. The history of metacognition research in humans is replete with apparent self-knowledge that, on closer inspection, reduced to shallow heuristics and confabulation \citep{nisbett1977telling, koriat1997monitoring}; our results suggest the same caution is warranted here.

Beyond these particular confounds, we have argued that success in current paradigms underdetermines introspection in the strong sense. Because that notion concerns \emph{how} a self-report is produced---by a state that represents something about a first-order representation, rather than by first-order computation alone---and because the two can yield identical reports, adjudicating between them requires designs on which they make divergent predictions and, ultimately, converging mechanistic evidence. This is not a counsel of despair. Whereas human metacognition research must infer mechanism from carefully constructed behavioral dissociations \citep{fleming2014measure, fleming2017self}, the internal computations of LLMs are fully observable, making converging evidence a realistic standard.

\label{sec:conclusions}

\section*{Acknowledgments}
This work was supported in part through the NYU IT High Performance Computing resources, services, and staff expertise

We thank Noam Steinmetz Yalon and Mor Geva for openly sharing the data used in \citet{steinmetz2026belief}, and for generously answering our questions about their methodology and helping us to refine our core claims. We also thank Dave Chalmers, John Morrison, Yanai Elazar, Yoav Goldberg, Jack Lindsey, Matt Mandelkern,  and Gal Vishne for their feedback. 

\bibliography{colm2026_conference}

@article{kadavath2022language,
  title={Language Models (Mostly) Know What They Know},
  author={Kadavath, Saurav and Conerly, Tom and Askell, Amanda and Henighan, Tom and Drain, Dawn and Perez, Ethan and Schiefer, Nicholas and Hatfield-Dodds, Zac and DasSarma, Nova and Tran-Johnson, Eli and others},
  journal={arXiv preprint arXiv:2207.05221},
  year={2022},
  url={https://arxiv.org/abs/2207.05221}
}

@article{lin2022teaching,
  title={Teaching Models to Express Their Uncertainty in Words},
  author={Lin, Stephanie and Hilton, Jacob and Evans, Owain},
  journal={Transactions on Machine Learning Research},
  year={2022},
  url={https://openreview.net/forum?id=8s8K2UZGTZ}
}

@inproceedings{yona2024faithful,
  title={Can Large Language Models Faithfully Express Their Intrinsic Uncertainty in Words?},
  author={Yona, Gal and Aharoni, Roee and Geva, Mor},
  booktitle={Proceedings of the 2024 Conference on Empirical Methods in Natural Language Processing},
  pages={7752--7764},
  year={2024},
  url={https://aclanthology.org/2024.emnlp-main.443/}
}

@inproceedings{burns2023discovering,
  title={Discovering Latent Knowledge in Language Models Without Supervision},
  author={Burns, Collin and Ye, Haotian and Klein, Dan and Steinhardt, Jacob},
  booktitle={The Eleventh International Conference on Learning Representations},
  year={2023},
  url={https://openreview.net/forum?id=ETKGuby0hcs}
}

@inproceedings{marks2024geometry,
  title={The Geometry of Truth: Emergent Linear Structure in Large Language Model Representations of True/False Datasets},
  author={Marks, Samuel and Tegmark, Max},
  booktitle={First Conference on Language Modeling},
  year={2024},
  url={https://openreview.net/forum?id=aajyHYjjsk}
}

@inproceedings{azaria2023internal,
  title={The Internal State of an {LLM} Knows When It's Lying},
  author={Azaria, Amos and Mitchell, Tom},
  booktitle={Findings of the Association for Computational Linguistics: EMNLP 2023},
  pages={967--976},
  year={2023},
  url={https://aclanthology.org/2023.findings-emnlp.68/}
}

@inproceedings{liu2023cognitive,
  title={Cognitive Dissonance: Why Do Language Model Outputs Disagree with Internal Representations of Truthfulness?},
  author={Liu, Kevin and Casper, Stephen and Hadfield-Menell, Dylan and Andreas, Jacob},
  booktitle={Proceedings of the 2023 Conference on Empirical Methods in Natural Language Processing},
  pages={4791--4797},
  year={2023},
  url={https://aclanthology.org/2023.emnlp-main.291/}
}

@article{jian2025metacognitive,
  title={Language Models Are Capable of Metacognitive Monitoring and Control of Their Internal Activations},
  author={Ji-An, Li and Xiong, Hua-Dong and Wilson, Robert and Mattar, Marcelo G and Benna, Marcus K},
  booktitle={The Thirty-ninth Annual Conference on Neural Information Processing Systems},
  year={2025},
  url={https://arxiv.org/abs/2505.13763}
}

@article{butlin2023consciousness,
  title={Consciousness in Artificial Intelligence: Insights from the Science of Consciousness},
  author={Butlin, Patrick and Long, Robert and Elmoznino, Eric and Bengio, Yoshua and Birch, Jonathan and Constant, Axel and Deane, George and Fleming, Stephen M and Frith, Chris and Ji, Xu and others},
  journal={arXiv preprint arXiv:2308.08708},
  year={2023},
  url={https://arxiv.org/abs/2308.08708}
}

@article{steinmetz2026belief,
  title={Indications of Belief-Guided Agency and Meta-Cognitive Monitoring in Large Language Models},
  author={Steinmetz Yalon, Noam and Goldstein, Ariel and Mudrik, Liad and Geva, Mor},
  journal={arXiv preprint arXiv:2602.02467},
  year={2026},
  url={https://arxiv.org/abs/2602.02467}
}

@article{shanahan2023role,
  title={Role Play with Large Language Models},
  author={Shanahan, Murray and McDonell, Kyle and Reynolds, Laria},
  journal={Nature},
  volume={623},
  number={7987},
  pages={493--498},
  year={2023},
  url={https://doi.org/10.1038/s41586-023-06647-8}
}

@article{turpin2023language,
  title={Language Models Don't Always Say What They Think: Unfaithful Explanations in Chain-of-Thought Prompting},
  author={Turpin, Miles and Michael, Julian and Perez, Ethan and Bowman, Samuel},
  journal={Advances in Neural Information Processing Systems},
  volume={36},
  pages={74952--74965},
  year={2023},
  url={https://arxiv.org/abs/2305.04388}
}

@inproceedings{kandpal2023large,
  title={Large Language Models Struggle to Learn Long-Tail Knowledge},
  author={Kandpal, Nikhil and Deng, Haikang and Roberts, Adam and Wallace, Eric and Raffel, Colin},
  booktitle={International Conference on Machine Learning},
  pages={15696--15707},
  year={2023},
  organization={PMLR},
  url={https://arxiv.org/abs/2211.08411}
}

@inproceedings{mallen2023trust,
  title={When Not to Trust Language Models: Investigating Effectiveness of Parametric and Non-Parametric Memories},
  author={Mallen, Alex and Asai, Akari and Zhong, Victor and Das, Rajarshi and Khashabi, Daniel and Hajishirzi, Hannaneh},
  booktitle={Proceedings of the 61st Annual Meeting of the Association for Computational Linguistics (Volume 1: Long Papers)},
  pages={9802--9822},
  year={2023},
  url={https://aclanthology.org/2023.acl-long.546/}
}

@inproceedings{xie2024adaptive,
  title={Adaptive Chameleon or Stubborn Sloth: Revealing the Behavior of Large Language Models in Knowledge Conflicts},
  author={Xie, Jian and Zhang, Kai and Chen, Jiangjie and Lou, Renze and Su, Yu},
  booktitle={The Twelfth International Conference on Learning Representations},
  year={2024},
  url={https://openreview.net/forum?id=auKAUJZMO6}
}

@inproceedings{
ravfogel2025emergence,
title={Emergence of Linear Truth Encodings in Language Models},
author={Shauli Ravfogel and Gilad Yehudai and Tal Linzen and Joan Bruna and Alberto Bietti},
booktitle={The Thirty-ninth Annual Conference on Neural Information Processing Systems},
year={2025},
url={https://openreview.net/forum?id=UQxUhFGUyk}
}

@inproceedings{slobodkin-etal-2023-curious,
    title = "The Curious Case of Hallucinatory (Un)answerability: Finding Truths in the Hidden States of Over-Confident Large Language Models",
    author = "Slobodkin, Aviv  and
      Goldman, Omer  and
      Caciularu, Avi  and
      Dagan, Ido  and
      Ravfogel, Shauli",
    editor = "Bouamor, Houda  and
      Pino, Juan  and
      Bali, Kalika",
    booktitle = "Proceedings of the 2023 Conference on Empirical Methods in Natural Language Processing",
    month = dec,
    year = "2023",
    address = "Singapore",
    publisher = "Association for Computational Linguistics",
    url = "https://aclanthology.org/2023.emnlp-main.220/",
    doi = "10.18653/v1/2023.emnlp-main.220",
    pages = "3607--3625"
}

@article{nisbett1977telling,
  title={Telling more than we can know: Verbal reports on mental processes.},
  author={Nisbett, Richard E and Wilson, Timothy D},
  journal={Psychological review},
  volume={84},
  number={3},
  pages={231},
  year={1977},
  publisher={American Psychological Association}
}

@article{flavell1979metacognition,
  title={Metacognition and cognitive monitoring: A new area of cognitive--developmental inquiry.},
  author={Flavell, John H},
  journal={American psychologist},
  volume={34},
  number={10},
  pages={906},
  year={1979},
  publisher={American Psychological Association}
}

@incollection{nelson1990metamemory,
  title={Metamemory: A theoretical framework and new findings},
  author={Nelson, Thomas O},
  booktitle={Psychology of learning and motivation},
  volume={26},
  pages={125--173},
  year={1990},
  publisher={Elsevier}
}

@article{fleming2014measure,
  author  = {Fleming, Stephen M. and Lau, Hakwan C.},
  title   = {How to measure metacognition},
  journal = {Frontiers in Human Neuroscience},
  volume  = {8},
  pages   = {443},
  year    = {2014},
  doi     = {10.3389/fnhum.2014.00443}
}

@article{lindsey2025emergent,
  author={Lindsey, Jack},
  title={Emergent Introspective Awareness in Large Language Models},
  journal={Transformer Circuits Thread},
  year={2025},
  url={https://transformer-circuits.pub/2025/introspection/index.html}
}

@article{binder2024looking,
  title={Looking Inward: Language Models Can Learn About Themselves by Introspection},
  author={Binder, Felix J and Chua, James and Korbak, Tomek and Sleight, Henry and Hughes, John and Long, Robert and Perez, Ethan and Turpin, Miles and Evans, Owain},
  journal={arXiv preprint arXiv:2410.13787},
  year={2024},
  url={https://arxiv.org/abs/2410.13787}
}

@article{song2025privileged,
  title={Privileged Self-Access Matters for Introspection in {AI}},
  author={Song, Siyuan and Lederman, Harvey and Hu, Jennifer and Mahowald, Kyle},
  journal={arXiv preprint arXiv:2508.14802},
  year={2025},
  url={https://arxiv.org/abs/2508.14802}
}

@article{koriat1997monitoring,
  title={Monitoring one's own knowledge during study: A cue-utilization approach to judgments of learning.},
  author={Koriat, Asher},
  journal={Journal of experimental psychology: General},
  volume={126},
  number={4},
  pages={349},
  year={1997},
  publisher={American Psychological Association}
}

@article{li2023inference,
  title={Inference-time intervention: Eliciting truthful answers from a language model},
  author={Li, Kenneth and Patel, Oam and Vi{\'e}gas, Fernanda and Pfister, Hanspeter and Wattenberg, Martin},
  journal={Advances in Neural Information Processing Systems},
  volume={36},
  pages={41451--41530},
  year={2023}
}

@inproceedings{singh2024representation,
  title={Representation surgery: theory and practice of affine steering},
  author={Singh, Shashwat and Ravfogel, Shauli and Herzig, Jonathan and Aharoni, Roee and Cotterell, Ryan and Kumaraguru, Ponnurangam},
  booktitle={Proceedings of the 41st International Conference on Machine Learning},
  pages={45663--45680},
  year={2024}
}

@inproceedings{ghandeharioun2024patchscopes,
  title={Patchscopes: A Unifying Framework for Inspecting Hidden Representations of Language Models},
  author={Ghandeharioun, Asma and Caciularu, Avi and Pearce, Adam and Dixon, Lucas and Geva, Mor},
  booktitle={Forty-first International Conference on Machine Learning},
  year={2024},
  url={https://arxiv.org/abs/2401.06102}
}

@article{karvonen2025activation,
  title={Activation oracles: Training and evaluating llms as general-purpose activation explainers},
  author={Karvonen, Adam and Chua, James and Dumas, Cl{\'e}ment and Fraser-Taliente, Kit and Kantamneni, Subhash and Minder, Julian and Ong, Euan and Sharma, Arnab Sen and Wen, Daniel and Evans, Owain and others},
  journal={arXiv preprint arXiv:2512.15674},
  year={2025}
}

@article{li2025training,
  title={Training language models to explain their own computations},
  author={Li, Belinda Z and Guo, Zifan Carl and Huang, Vincent and Steinhardt, Jacob and Andreas, Jacob},
  journal={arXiv preprint arXiv:2511.08579},
  year={2025}
}

@misc{vogel2025small,
  title={Small Models Can Introspect, Too},
  author={Vogel, Theia},
  howpublished={\url{https://vgel.me/posts/qwen-introspection/}},
  year={2025}
}

@misc{rivera2026steeringawarenessdetectingactivation,
      title={Steering Awareness: Detecting Activation Steering from Within}, 
      author={Joshua Fonseca Rivera and David Demitri Africa},
      year={2026},
      eprint={2511.21399},
      archivePrefix={arXiv},
      primaryClass={cs.CL},
      url={https://arxiv.org/abs/2511.21399}, 
}

@misc{hendrycks2023aligningaisharedhuman,
      title={Aligning AI With Shared Human Values}, 
      author={Dan Hendrycks and Collin Burns and Steven Basart and Andrew Critch and Jerry Li and Dawn Song and Jacob Steinhardt},
      year={2023},
      eprint={2008.02275},
      archivePrefix={arXiv},
      primaryClass={cs.CY},
      url={https://arxiv.org/abs/2008.02275}, 
}

@article{meng2022locating,
  title={Locating and editing factual associations in gpt},
  author={Meng, Kevin and Bau, David and Andonian, Alex and Belinkov, Yonatan},
  journal={Advances in neural information processing systems},
  volume={35},
  pages={17359--17372},
  year={2022}
}

@misc{grattafiori2024llama3herdmodels,
      title={The Llama 3 Herd of Models}, 
      author={Aaron Grattafiori and Abhimanyu Dubey and Abhinav Jauhri and Abhinav Pandey and Abhishek Kadian and Ahmad Al-Dahle and Aiesha Letman and Akhil Mathur and Alan Schelten and Alex Vaughan and Amy Yang and Angela Fan and Anirudh Goyal and Anthony Hartshorn and Aobo Yang and Archi Mitra and Archie Sravankumar and Artem Korenev and Arthur Hinsvark and Arun Rao and Aston Zhang and Aurelien Rodriguez and Austen Gregerson and Ava Spataru and Baptiste Roziere and Bethany Biron and Binh Tang and Bobbie Chern and Charlotte Caucheteux and Chaya Nayak and Chloe Bi and Chris Marra and Chris McConnell and Christian Keller and Christophe Touret and Chunyang Wu and Corinne Wong and Cristian Canton Ferrer and Cyrus Nikolaidis and Damien Allonsius and Daniel Song and Danielle Pintz and Danny Livshits and Danny Wyatt and David Esiobu and Dhruv Choudhary and Dhruv Mahajan and Diego Garcia-Olano and Diego Perino and Dieuwke Hupkes and Egor Lakomkin and Ehab AlBadawy and Elina Lobanova and Emily Dinan and Eric Michael Smith and Filip Radenovic and Francisco Guzmán and Frank Zhang and Gabriel Synnaeve and Gabrielle Lee and Georgia Lewis Anderson and Govind Thattai and Graeme Nail and Gregoire Mialon and Guan Pang and Guillem Cucurell and Hailey Nguyen and Hannah Korevaar and Hu Xu and Hugo Touvron and Iliyan Zarov and Imanol Arrieta Ibarra and Isabel Kloumann and Ishan Misra and Ivan Evtimov and Jack Zhang and Jade Copet and Jaewon Lee and Jan Geffert and Jana Vranes and Jason Park and Jay Mahadeokar and Jeet Shah and Jelmer van der Linde and Jennifer Billock and Jenny Hong and Jenya Lee and Jeremy Fu and Jianfeng Chi and Jianyu Huang and Jiawen Liu and Jie Wang and Jiecao Yu and Joanna Bitton and Joe Spisak and Jongsoo Park and Joseph Rocca and Joshua Johnstun and Joshua Saxe and Junteng Jia and Kalyan Vasuden Alwala and Karthik Prasad and Kartikeya Upasani and Kate Plawiak and Ke Li and Kenneth Heafield and Kevin Stone and Khalid El-Arini and Krithika Iyer and Kshitiz Malik and Kuenley Chiu and Kunal Bhalla and Kushal Lakhotia and Lauren Rantala-Yeary and Laurens van der Maaten and Lawrence Chen and Liang Tan and Liz Jenkins and Louis Martin and Lovish Madaan and Lubo Malo and Lukas Blecher and Lukas Landzaat and Luke de Oliveira and Madeline Muzzi and Mahesh Pasupuleti and Mannat Singh and Manohar Paluri and Marcin Kardas and Maria Tsimpoukelli and Mathew Oldham and Mathieu Rita and Maya Pavlova and Melanie Kambadur and Mike Lewis and Min Si and Mitesh Kumar Singh and Mona Hassan and Naman Goyal and Narjes Torabi and Nikolay Bashlykov and Nikolay Bogoychev and Niladri Chatterji and Ning Zhang and Olivier Duchenne and Onur Çelebi and Patrick Alrassy and Pengchuan Zhang and Pengwei Li and Petar Vasic and Peter Weng and Prajjwal Bhargava and Pratik Dubal and Praveen Krishnan and Punit Singh Koura and Puxin Xu and Qing He and Qingxiao Dong and Ragavan Srinivasan and Raj Ganapathy and Ramon Calderer and Ricardo Silveira Cabral and Robert Stojnic and Roberta Raileanu and Rohan Maheswari and Rohit Girdhar and Rohit Patel and Romain Sauvestre and Ronnie Polidoro and Roshan Sumbaly and Ross Taylor and Ruan Silva and Rui Hou and Rui Wang and Saghar Hosseini and Sahana Chennabasappa and Sanjay Singh and Sean Bell and Seohyun Sonia Kim and Sergey Edunov and Shaoliang Nie and Sharan Narang and Sharath Raparthy and Sheng Shen and Shengye Wan and Shruti Bhosale and Shun Zhang and Simon Vandenhende and Soumya Batra and Spencer Whitman and Sten Sootla and Stephane Collot and Suchin Gururangan and Sydney Borodinsky and Tamar Herman and Tara Fowler and Tarek Sheasha and Thomas Georgiou and Thomas Scialom and Tobias Speckbacher and Todor Mihaylov and Tong Xiao and Ujjwal Karn and Vedanuj Goswami and Vibhor Gupta and Vignesh Ramanathan and Viktor Kerkez and Vincent Gonguet and Virginie Do and Vish Vogeti and Vítor Albiero and Vladan Petrovic and Weiwei Chu and Wenhan Xiong and Wenyin Fu and Whitney Meers and Xavier Martinet and Xiaodong Wang and Xiaofang Wang and Xiaoqing Ellen Tan and Xide Xia and Xinfeng Xie and Xuchao Jia and Xuewei Wang and Yaelle Goldschlag and Yashesh Gaur and Yasmine Babaei and Yi Wen and Yiwen Song and Yuchen Zhang and Yue Li and Yuning Mao and Zacharie Delpierre Coudert and Zheng Yan and Zhengxing Chen and Zoe Papakipos and Aaditya Singh and Aayushi Srivastava and Abha Jain and Adam Kelsey and Adam Shajnfeld and Adithya Gangidi and Adolfo Victoria and Ahuva Goldstand and Ajay Menon and Ajay Sharma and Alex Boesenberg and Alexei Baevski and Allie Feinstein and Amanda Kallet and Amit Sangani and Amos Teo and Anam Yunus and Andrei Lupu and Andres Alvarado and Andrew Caples and Andrew Gu and Andrew Ho and Andrew Poulton and Andrew Ryan and Ankit Ramchandani and Annie Dong and Annie Franco and Anuj Goyal and Aparajita Saraf and Arkabandhu Chowdhury and Ashley Gabriel and Ashwin Bharambe and Assaf Eisenman and Azadeh Yazdan and Beau James and Ben Maurer and Benjamin Leonhardi and Bernie Huang and Beth Loyd and Beto De Paola and Bhargavi Paranjape and Bing Liu and Bo Wu and Boyu Ni and Braden Hancock and Bram Wasti and Brandon Spence and Brani Stojkovic and Brian Gamido and Britt Montalvo and Carl Parker and Carly Burton and Catalina Mejia and Ce Liu and Changhan Wang and Changkyu Kim and Chao Zhou and Chester Hu and Ching-Hsiang Chu and Chris Cai and Chris Tindal and Christoph Feichtenhofer and Cynthia Gao and Damon Civin and Dana Beaty and Daniel Kreymer and Daniel Li and David Adkins and David Xu and Davide Testuggine and Delia David and Devi Parikh and Diana Liskovich and Didem Foss and Dingkang Wang and Duc Le and Dustin Holland and Edward Dowling and Eissa Jamil and Elaine Montgomery and Eleonora Presani and Emily Hahn and Emily Wood and Eric-Tuan Le and Erik Brinkman and Esteban Arcaute and Evan Dunbar and Evan Smothers and Fei Sun and Felix Kreuk and Feng Tian and Filippos Kokkinos and Firat Ozgenel and Francesco Caggioni and Frank Kanayet and Frank Seide and Gabriela Medina Florez and Gabriella Schwarz and Gada Badeer and Georgia Swee and Gil Halpern and Grant Herman and Grigory Sizov and Guangyi and Zhang and Guna Lakshminarayanan and Hakan Inan and Hamid Shojanazeri and Han Zou and Hannah Wang and Hanwen Zha and Haroun Habeeb and Harrison Rudolph and Helen Suk and Henry Aspegren and Hunter Goldman and Hongyuan Zhan and Ibrahim Damlaj and Igor Molybog and Igor Tufanov and Ilias Leontiadis and Irina-Elena Veliche and Itai Gat and Jake Weissman and James Geboski and James Kohli and Janice Lam and Japhet Asher and Jean-Baptiste Gaya and Jeff Marcus and Jeff Tang and Jennifer Chan and Jenny Zhen and Jeremy Reizenstein and Jeremy Teboul and Jessica Zhong and Jian Jin and Jingyi Yang and Joe Cummings and Jon Carvill and Jon Shepard and Jonathan McPhie and Jonathan Torres and Josh Ginsburg and Junjie Wang and Kai Wu and Kam Hou U and Karan Saxena and Kartikay Khandelwal and Katayoun Zand and Kathy Matosich and Kaushik Veeraraghavan and Kelly Michelena and Keqian Li and Kiran Jagadeesh and Kun Huang and Kunal Chawla and Kyle Huang and Lailin Chen and Lakshya Garg and Lavender A and Leandro Silva and Lee Bell and Lei Zhang and Liangpeng Guo and Licheng Yu and Liron Moshkovich and Luca Wehrstedt and Madian Khabsa and Manav Avalani and Manish Bhatt and Martynas Mankus and Matan Hasson and Matthew Lennie and Matthias Reso and Maxim Groshev and Maxim Naumov and Maya Lathi and Meghan Keneally and Miao Liu and Michael L. Seltzer and Michal Valko and Michelle Restrepo and Mihir Patel and Mik Vyatskov and Mikayel Samvelyan and Mike Clark and Mike Macey and Mike Wang and Miquel Jubert Hermoso and Mo Metanat and Mohammad Rastegari and Munish Bansal and Nandhini Santhanam and Natascha Parks and Natasha White and Navyata Bawa and Nayan Singhal and Nick Egebo and Nicolas Usunier and Nikhil Mehta and Nikolay Pavlovich Laptev and Ning Dong and Norman Cheng and Oleg Chernoguz and Olivia Hart and Omkar Salpekar and Ozlem Kalinli and Parkin Kent and Parth Parekh and Paul Saab and Pavan Balaji and Pedro Rittner and Philip Bontrager and Pierre Roux and Piotr Dollar and Polina Zvyagina and Prashant Ratanchandani and Pritish Yuvraj and Qian Liang and Rachad Alao and Rachel Rodriguez and Rafi Ayub and Raghotham Murthy and Raghu Nayani and Rahul Mitra and Rangaprabhu Parthasarathy and Raymond Li and Rebekkah Hogan and Robin Battey and Rocky Wang and Russ Howes and Ruty Rinott and Sachin Mehta and Sachin Siby and Sai Jayesh Bondu and Samyak Datta and Sara Chugh and Sara Hunt and Sargun Dhillon and Sasha Sidorov and Satadru Pan and Saurabh Mahajan and Saurabh Verma and Seiji Yamamoto and Sharadh Ramaswamy and Shaun Lindsay and Shaun Lindsay and Sheng Feng and Shenghao Lin and Shengxin Cindy Zha and Shishir Patil and Shiva Shankar and Shuqiang Zhang and Shuqiang Zhang and Sinong Wang and Sneha Agarwal and Soji Sajuyigbe and Soumith Chintala and Stephanie Max and Stephen Chen and Steve Kehoe and Steve Satterfield and Sudarshan Govindaprasad and Sumit Gupta and Summer Deng and Sungmin Cho and Sunny Virk and Suraj Subramanian and Sy Choudhury and Sydney Goldman and Tal Remez and Tamar Glaser and Tamara Best and Thilo Koehler and Thomas Robinson and Tianhe Li and Tianjun Zhang and Tim Matthews and Timothy Chou and Tzook Shaked and Varun Vontimitta and Victoria Ajayi and Victoria Montanez and Vijai Mohan and Vinay Satish Kumar and Vishal Mangla and Vlad Ionescu and Vlad Poenaru and Vlad Tiberiu Mihailescu and Vladimir Ivanov and Wei Li and Wenchen Wang and Wenwen Jiang and Wes Bouaziz and Will Constable and Xiaocheng Tang and Xiaojian Wu and Xiaolan Wang and Xilun Wu and Xinbo Gao and Yaniv Kleinman and Yanjun Chen and Ye Hu and Ye Jia and Ye Qi and Yenda Li and Yilin Zhang and Ying Zhang and Yossi Adi and Youngjin Nam and Yu and Wang and Yu Zhao and Yuchen Hao and Yundi Qian and Yunlu Li and Yuzi He and Zach Rait and Zachary DeVito and Zef Rosnbrick and Zhaoduo Wen and Zhenyu Yang and Zhiwei Zhao and Zhiyu Ma},
      year={2024},
      eprint={2407.21783},
      archivePrefix={arXiv},
      primaryClass={cs.AI},
      url={https://arxiv.org/abs/2407.21783}, 
}

@misc{qwen2025qwen25technicalreport,
      title={Qwen2.5 Technical Report}, 
      author={Qwen and : and An Yang and Baosong Yang and Beichen Zhang and Binyuan Hui and Bo Zheng and Bowen Yu and Chengyuan Li and Dayiheng Liu and Fei Huang and Haoran Wei and Huan Lin and Jian Yang and Jianhong Tu and Jianwei Zhang and Jianxin Yang and Jiaxi Yang and Jingren Zhou and Junyang Lin and Kai Dang and Keming Lu and Keqin Bao and Kexin Yang and Le Yu and Mei Li and Mingfeng Xue and Pei Zhang and Qin Zhu and Rui Men and Runji Lin and Tianhao Li and Tianyi Tang and Tingyu Xia and Xingzhang Ren and Xuancheng Ren and Yang Fan and Yang Su and Yichang Zhang and Yu Wan and Yuqiong Liu and Zeyu Cui and Zhenru Zhang and Zihan Qiu},
      year={2025},
      eprint={2412.15115},
      archivePrefix={arXiv},
      primaryClass={cs.CL},
      url={https://arxiv.org/abs/2412.15115}, 
}

@inproceedings{macar2026mechanisms,
  title={Mechanisms of Introspective Awareness},
  author={Macar, Uzay and Yang, Li and Wang, Atticus and Wallich, Peter and Ameisen, Emmanuel and Lindsey, Jack},
  booktitle={ICLR 2026 Workshop-From Human Cognition to AI Reasoning: Models, Methods, and Applications},
  year={2026}
}

@article{lederman2026dissociating,
  title={Dissociating Direct Access from Inference in AI Introspection},
  author={Lederman, Harvey and Mahowald, Kyle},
  journal={arXiv e-prints},
  pages={arXiv--2603},
  year={2026}
}

@book{armstrong1968,
  author    = {Armstrong, D. M.},
  title     = {A Materialist Theory of the Mind},
  publisher = {Routledge \& Kegan Paul},
  address   = {London},
  year      = {1968}
}

@book{rosenthal2005consciousness,
  title={Consciousness and mind},
  author={Rosenthal, David},
  year={2005},
  publisher={Clarendon Press}
}

@book{nichols2003mindreading,
  title={Mindreading: An integrated account of pretence, self-awareness, and understanding other minds},
  author={Nichols, Shaun and Stich, Stephen P},
  year={2003},
  publisher={Oxford University Press}
}

@book{carruthers2011opacity,
  title={The opacity of mind: An integrative theory of self-knowledge},
  author={Carruthers, Peter},
  year={2011},
  publisher={OUP Oxford}
}

@book{byrne2018transparency,
  title={Transparency and self-knowledge},
  author={Byrne, Alex},
  year={2018},
  publisher={Oxford University Press}
}

@misc{gemmateam2025gemma3technicalreport,
      title={Gemma 3 Technical Report}, 
      author={Gemma Team and Aishwarya Kamath and Johan Ferret and Shreya Pathak and Nino Vieillard and Ramona Merhej and Sarah Perrin and Tatiana Matejovicova and Alexandre Ramé and Morgane Rivière and Louis Rouillard and Thomas Mesnard and Geoffrey Cideron and Jean-bastien Grill and Sabela Ramos and Edouard Yvinec and Michelle Casbon and Etienne Pot and Ivo Penchev and Gaël Liu and Francesco Visin and Kathleen Kenealy and Lucas Beyer and Xiaohai Zhai and Anton Tsitsulin and Robert Busa-Fekete and Alex Feng and Noveen Sachdeva and Benjamin Coleman and Yi Gao and Basil Mustafa and Iain Barr and Emilio Parisotto and David Tian and Matan Eyal and Colin Cherry and Jan-Thorsten Peter and Danila Sinopalnikov and Surya Bhupatiraju and Rishabh Agarwal and Mehran Kazemi and Dan Malkin and Ravin Kumar and David Vilar and Idan Brusilovsky and Jiaming Luo and Andreas Steiner and Abe Friesen and Abhanshu Sharma and Abheesht Sharma and Adi Mayrav Gilady and Adrian Goedeckemeyer and Alaa Saade and Alex Feng and Alexander Kolesnikov and Alexei Bendebury and Alvin Abdagic and Amit Vadi and András György and André Susano Pinto and Anil Das and Ankur Bapna and Antoine Miech and Antoine Yang and Antonia Paterson and Ashish Shenoy and Ayan Chakrabarti and Bilal Piot and Bo Wu and Bobak Shahriari and Bryce Petrini and Charlie Chen and Charline Le Lan and Christopher A. Choquette-Choo and CJ Carey and Cormac Brick and Daniel Deutsch and Danielle Eisenbud and Dee Cattle and Derek Cheng and Dimitris Paparas and Divyashree Shivakumar Sreepathihalli and Doug Reid and Dustin Tran and Dustin Zelle and Eric Noland and Erwin Huizenga and Eugene Kharitonov and Frederick Liu and Gagik Amirkhanyan and Glenn Cameron and Hadi Hashemi and Hanna Klimczak-Plucińska and Harman Singh and Harsh Mehta and Harshal Tushar Lehri and Hussein Hazimeh and Ian Ballantyne and Idan Szpektor and Ivan Nardini and Jean Pouget-Abadie and Jetha Chan and Joe Stanton and John Wieting and Jonathan Lai and Jordi Orbay and Joseph Fernandez and Josh Newlan and Ju-yeong Ji and Jyotinder Singh and Kat Black and Kathy Yu and Kevin Hui and Kiran Vodrahalli and Klaus Greff and Linhai Qiu and Marcella Valentine and Marina Coelho and Marvin Ritter and Matt Hoffman and Matthew Watson and Mayank Chaturvedi and Michael Moynihan and Min Ma and Nabila Babar and Natasha Noy and Nathan Byrd and Nick Roy and Nikola Momchev and Nilay Chauhan and Noveen Sachdeva and Oskar Bunyan and Pankil Botarda and Paul Caron and Paul Kishan Rubenstein and Phil Culliton and Philipp Schmid and Pier Giuseppe Sessa and Pingmei Xu and Piotr Stanczyk and Pouya Tafti and Rakesh Shivanna and Renjie Wu and Renke Pan and Reza Rokni and Rob Willoughby and Rohith Vallu and Ryan Mullins and Sammy Jerome and Sara Smoot and Sertan Girgin and Shariq Iqbal and Shashir Reddy and Shruti Sheth and Siim Põder and Sijal Bhatnagar and Sindhu Raghuram Panyam and Sivan Eiger and Susan Zhang and Tianqi Liu and Trevor Yacovone and Tyler Liechty and Uday Kalra and Utku Evci and Vedant Misra and Vincent Roseberry and Vlad Feinberg and Vlad Kolesnikov and Woohyun Han and Woosuk Kwon and Xi Chen and Yinlam Chow and Yuvein Zhu and Zichuan Wei and Zoltan Egyed and Victor Cotruta and Minh Giang and Phoebe Kirk and Anand Rao and Kat Black and Nabila Babar and Jessica Lo and Erica Moreira and Luiz Gustavo Martins and Omar Sanseviero and Lucas Gonzalez and Zach Gleicher and Tris Warkentin and Vahab Mirrokni and Evan Senter and Eli Collins and Joelle Barral and Zoubin Ghahramani and Raia Hadsell and Yossi Matias and D. Sculley and Slav Petrov and Noah Fiedel and Noam Shazeer and Oriol Vinyals and Jeff Dean and Demis Hassabis and Koray Kavukcuoglu and Clement Farabet and Elena Buchatskaya and Jean-Baptiste Alayrac and Rohan Anil and Dmitry and Lepikhin and Sebastian Borgeaud and Olivier Bachem and Armand Joulin and Alek Andreev and Cassidy Hardin and Robert Dadashi and Léonard Hussenot},
      year={2025},
      eprint={2503.19786},
      archivePrefix={arXiv},
      primaryClass={cs.CL},
      url={https://arxiv.org/abs/2503.19786}, 
}

@article{maniscalco2012signal,
  title   = {A signal detection theoretic approach for estimating metacognitive
             sensitivity from confidence ratings},
  author  = {Maniscalco, Brian and Lau, Hakwan},
  journal = {Consciousness and Cognition},
  volume  = {21},
  number  = {1},
  pages   = {422--430},
  year    = {2012},
  doi     = {10.1016/j.concog.2011.09.021}
}

@article{fleming2017self,
  title   = {Self-evaluation of decision-making: A general {B}ayesian framework
             for metacognitive computation},
  author  = {Fleming, Stephen M. and Daw, Nathaniel D.},
  journal = {Psychological Review},
  volume  = {124},
  number  = {1},
  pages   = {91--114},
  year    = {2017},
  doi     = {10.1037/rev0000045}
}

@article{smith2003comparative,
  title   = {The comparative psychology of uncertainty monitoring and
             metacognition},
  author  = {Smith, J. David and Shields, Wendy E. and Washburn, David A.},
  journal = {Behavioral and Brain Sciences},
  volume  = {26},
  number  = {3},
  pages   = {317--339},
  year    = {2003}
}

@article{hampton2001rhesus,
  title   = {Rhesus monkeys know when they remember},
  author  = {Hampton, Robert R.},
  journal = {Proceedings of the National Academy of Sciences},
  volume  = {98},
  number  = {9},
  pages   = {5359--5362},
  year    = {2001},
  doi     = {10.1073/pnas.071600998}
}

@article{carruthers2008meta,
  title   = {Meta-cognition in animals: A skeptical look},
  author  = {Carruthers, Peter},
  journal = {Mind \& Language},
  volume  = {23},
  number  = {1},
  pages   = {58--89},
  year    = {2008},
  doi     = {10.1111/j.1468-0017.2007.00329.x}
}

@article{kepecs2008neural,
  title   = {Neural correlates, computation and behavioural impact of decision
             confidence},
  author  = {Kepecs, Adam and Uchida, Naoshige and Zariwala, Hatim A. and
             Mainen, Zachary F.},
  journal = {Nature},
  volume  = {455},
  number  = {7210},
  pages   = {227--231},
  year    = {2008},
  doi     = {10.1038/nature07200}
}

@misc{plunkett2025self,
  title         = {Self-Interpretability: {LLM}s Can Describe Complex Internal
                   Processes that Drive Their Decisions},
  author        = {Plunkett, Dillon and Morris, Adam and Reddy, Keerthi and
                   Morales, Jorge},
  year          = {2025},
  eprint        = {2505.17120},
  archivePrefix = {arXiv},
  primaryClass  = {cs.CL},
  url           = {https://arxiv.org/abs/2505.17120}
}

@book{lycan1996,
  title     = {Consciousness and Experience},
  author    = {Lycan, William G.},
  publisher = {Bradford Books / MIT Press},
  address   = {Cambridge, MA},
  year      = {1996}
}

@incollection{dretske1986misrepresentation,
  title     = {Misrepresentation},
  author    = {Dretske, Fred},
  booktitle = {Belief: Form, Content, and Function},
  editor    = {Bogdan, Radu J.},
  publisher = {Oxford University Press},
  address   = {New York},
  pages     = {17--36},
  year      = {1986}
}

@article{millikan1989biosemantics,
  title   = {Biosemantics},
  author  = {Millikan, Ruth Garrett},
  journal = {Journal of Philosophy},
  volume  = {86},
  number  = {6},
  pages   = {281--297},
  year    = {1989}
}

@book{shea2018representation,
  title     = {Representation in Cognitive Science},
  author    = {Shea, Nicholas},
  publisher = {Oxford University Press},
  address   = {Oxford},
  year      = {2018}
}

@book{ramsey2007representation,
  title     = {Representation Reconsidered},
  author    = {Ramsey, William M.},
  publisher = {Cambridge University Press},
  address   = {Cambridge},
  year      = {2007},
  doi       = {10.1017/CBO9780511597954}
}

@misc{yang2025qwen3technicalreport,
      title={Qwen3 Technical Report}, 
      author={An Yang and Anfeng Li and Baosong Yang and Beichen Zhang and Binyuan Hui and Bo Zheng and Bowen Yu and Chang Gao and Chengen Huang and Chenxu Lv and Chujie Zheng and Dayiheng Liu and Fan Zhou and Fei Huang and Feng Hu and Hao Ge and Haoran Wei and Huan Lin and Jialong Tang and Jian Yang and Jianhong Tu and Jianwei Zhang and Jianxin Yang and Jiaxi Yang and Jing Zhou and Jingren Zhou and Junyang Lin and Kai Dang and Keqin Bao and Kexin Yang and Le Yu and Lianghao Deng and Mei Li and Mingfeng Xue and Mingze Li and Pei Zhang and Peng Wang and Qin Zhu and Rui Men and Ruize Gao and Shixuan Liu and Shuang Luo and Tianhao Li and Tianyi Tang and Wenbiao Yin and Xingzhang Ren and Xinyu Wang and Xinyu Zhang and Xuancheng Ren and Yang Fan and Yang Su and Yichang Zhang and Yinger Zhang and Yu Wan and Yuqiong Liu and Zekun Wang and Zeyu Cui and Zhenru Zhang and Zhipeng Zhou and Zihan Qiu},
      year={2025},
      eprint={2505.09388},
      archivePrefix={arXiv},
      primaryClass={cs.CL},
      url={https://arxiv.org/abs/2505.09388}, 
}
\bibliographystyle{colm2026_conference}

\newpage
\clearpage
\appendix
\section*{Appendix}
\crefalias{section}{appendix}

\section{Limitations}\label{app:limitations}

Our central conceptual claim is that introspection requires second-order computation, but we do not offer a criterion that would let one test for it directly, in part because the literature disagrees about what makes a state a representation \emph{of} another state at all (\cref{app:meta-representation}). Consequently, our experiments on the \citet{lindsey2025emergent} paradigm test a strictly weaker condition---whether reports track interventions on hidden states specifically, rather than irregularity in general---and a model that passed our three-way test would still not thereby be shown to deploy meta-representations. We discuss properties that would constitute stronger evidence in \cref{app:meta-representation}.

For the self-report classification paradigms, we show that a linear probe on uncontextualized embeddings matches the models' in-context accuracy, which suffices to show that the reported results do not require privileged access. It does not show that no version of these paradigms could satisfy the privileged-access condition.

\section{AI Use}
Large language models were used to assist with running and analyzing experiments, 
and with improving the clarity and presentation of the writing.

\section{Extended Related Work}
\label{app:related}

\subsection{Approaches to LLM metacognition}
\label{app:related-approaches}

The question of whether LLMs possess metacognitive abilities has been approached from several angles. One line of work investigates \emph{verbal calibration}, asking whether models express well-calibrated uncertainty about their answers \citep{kadavath2022language, lin2022teaching, yona2024faithful}. A second employs \emph{probing-based approaches} that extract internal representations of confidence or truthfulness from hidden states \citep{burns2023discovering, marks2024geometry, azaria2023internal, liu2023cognitive, slobodkin-etal-2023-curious, ravfogel2025emergence}. A third adopts \emph{neuroscience-inspired paradigms} that evaluate indicators of consciousness derived from cognitive theories \citep{butlin2023consciousness, steinmetz2026belief} or test whether models can report their own activation patterns \citep{jian2025metacognitive}.

\subsection{Controlled evaluations of self-report}
\label{app:related-controlled}

Three recent studies explicitly try to rule out deflationary explanations of successful self-report. We describe them in terms of the two conditions of \cref{sec:intro}: privileged access (i) and second-order computation (ii).

\paragraph{Predicting one's own behavior.} \citet{binder2024looking} operationalize introspection as a predictive advantage over an equally informed external model: a model introspects if it predicts properties of its own outputs better than another model can, given the same prompt. Their finetuned models show such an advantage. As the authors note, however, a model could succeed by \emph{self-simulation}---internally computing the response it would give, then reading off the queried property. Self-simulation satisfies condition (i), since the simulated response is not recoverable from the prompt by an external observer, while requiring nothing second-order: the queried property is extracted from a first-order computation rather than from a representation of one.

\paragraph{Reporting quantitative features of one's own decision process.} \citet{plunkett2025self} report the most carefully controlled positive result in this literature. They finetuned GPT-4o and GPT-4o-mini to make binary choices on behalf of $100$ hypothetical agents, each with its own choice domain (condos, refrigerators, vacations, etc.) defined by five quantitative attributes. Per agent, five attribute weights were sampled uniformly from $[-100,100]$ and the model was finetuned on $5{,}000$ example choices generated from them; the weights themselves never appeared in the training data. Logistic regression over held-out choices confirmed that the models had internalized the target weights ($r = .84$ and $.87$). When asked to report how heavily they weighted each attribute, the models' reports correlated with the weights actually driving their choices ($r = .54$ and $.50$). A second round of finetuning, on examples of accurate self-report, raised this to $r = .74$ and $.75$ on held-out domains, and also improved reporting of the models' \emph{native} weights in domains absent from all finetuning ($.46 \rightarrow .71$ and $.40 \rightarrow .70$), indicating that the training generalizes.

The design controls well for condition (i). Because the target weights are random, common-sense priors about human preferences cannot be used to guess them---verified directly by prompting off-the-shelf models, whose reports are uncorrelated with the finetuned models' weights ($r = .10$ and $-.01$)---and because choices and reports are elicited in separate context windows, the models cannot infer their weights from their own outputs.\footnote{The weights are nonetheless estimable from behavior, which is how the authors measure them; privileged access here holds relative to an observer confined to a single context, but not to one permitted to sample the model's choices---the reference class \citet{song2025privileged} adopt.} Condition (ii) remains open, as the authors themselves note: finetuning could have updated the models' stored knowledge \emph{about} their preferences, so that accurate reports reflect retrieval of a self-descriptive fact rather than monitoring of the process generating a decision. They also observe that the paradigm probes stable, trait-level properties rather than the computation underlying any single forward pass, and explicitly disclaim the conscious-awareness reading of ``introspection.''

\paragraph{Privileged access as a computational-cost advantage.} \citet{song2025privileged} argue that comparison to a particular external model is too weak a criterion, and require instead a reliability advantage over \emph{any} process of equal or lower computational cost available to a third party; they show that apparent introspective successes can fail this test. Our empirical results can be read as further instances: in the paradigms we study, a linear probe over input embeddings matches the model's own performance.

\subsection{Verbalizing internal states}
\label{app:related-verbalizing}

A separate line of work trains models to verbalize information about their own activations: \citet{ghandeharioun2024patchscopes} patch hidden representations into prompts designed to extract information; \citet{karvonen2025activation} train ``Activation Oracles'' that answer questions about activation vectors; and \citet{li2025training} finetune models to describe their internal features and causal structures. These studies conclude that models exhibit privileged access, explaining their own internals better than other models can. This pattern, however, follows from the architecture: a model's forward pass operates in its own representational space by construction, whereas cross-model explanation requires additional alignment. It does not imply a distinct processing mode, and hence is better understood as an architectural consequence than as evidence for introspection in the psychological sense.

\subsection{Replications and critiques of concept-injection detection}
\label{app:related-replications}

\citet{vogel2025small} replicate \citet{lindsey2025emergent} in Qwen2.5-Coder-32B with appropriate prompting, and \citet{rivera2026steeringawarenessdetectingactivation} report $95.5\%$ detection with zero false positives in a Qwen2.5-32B model finetuned explicitly for steering awareness. \citet{lederman2026dissociating} offer the critique closest to ours, arguing that injection detection is content-agnostic: models detect that an anomaly occurred but cannot reliably identify the injected concept, often defaulting to high-frequency guesses. The two critiques target different links in the same chain: they dissociate detection from identification, whereas we grant accurate detection and ask what it consists in---a second-order representation of the model's states, or a first-order sensitivity to irregularity that is subsequently reported.

\subsection{Scope: pretrained versus finetuned models}
\label{app:related-scope}

Several studies above rely on finetuning: \citet{binder2024looking} train models to predict their own behavior, \citet{plunkett2025self} instill preferences and then improve self-report by finetuning, and \citet{rivera2026steeringawarenessdetectingactivation} train explicitly for steering awareness. Such designs establish what self-report capacities can be \emph{induced}. We ask instead whether introspection is present in pretrained models, since finetuning may solve the task by installing a task-specific mechanism rather than revealing a general capacity.

\section{What Would Make a State About the Model's Own Processing?}
\label{app:meta-representation}

In \cref{sec:construct-validity}, we argue that an accurate report about the model's internal state does not by itself establish meta-representation, which is a marker of introspection. For example, suppose that steering increases the norm of a hidden state and that a high-norm feature directly raises the probability of the sentence ``my activations were altered.'' This mechanism is sensitive to the model's own processing and may produce a reliable self-description, but its behavior is also explained by a feature detector that is causally connected to a given positive response. The reporting task alone does not tell us whether the detector represents the hidden state \emph{as} a hidden state.

We do not attempt to settle what representation ultimately requires. Different accounts disagree about whether representational content is fixed by information, function, structural correspondence, downstream use, or some combination of these \citep{dretske1986misrepresentation,millikan1989biosemantics,ramsey2007representation,shea2018representation}. Rather, we say that if self-report can be explained by a feature detection circuit that is not different in any significant way from the ones involved in ``non-introspective'' behavior, there is no reason for us to attribute a special ``introspective'' quality to the model. This also clarifies our use of ``second-order.'' The relevant sense is not merely that a computation takes another hidden state as input---as every transformer layer does that---but that its content concerns another state or process.

The distinction is closely related to \citet{ramsey2007representation}'s ``job description challenge.'' Calling a state a representation should explain something that is not already explained by describing it as a reliable causal relay. A bimetallic strip covaries with temperature, but describing it as a representation adds little unless the system uses it as a stand-in for temperature. Likewise, a feature that triggers a self-describing sentence may be a component of a monitor, but covariation and accurate reporting alone do not establish that role.

Several properties would provide stronger, though still defeasible, evidence. First, the feature should generalize in a manner appropriate to its proposed content: a representation of “my processing is abnormal” should generalize across relevant kinds of abnormality, whereas a representation of a specific layer property may properly be narrower. Second, downstream processes should use the feature as information about the model’s processing—for example, for calibration, hedging, or control. Finally, the proposed content should be distinguishable from the cue that activates the feature. If a norm detector is claimed to represent an intervention, for example, cases of high norm without an intervention and intervention without high norm can test whether it tracks the intervention itself or merely the correlated norm.

\section{Concepts}\label{app:concepts}

For our steering and gaslight experiments, we use the following set of ``human-interpretable'' concepts:

apple, astronomy, democracy, sushi, football, rivers, algorithms, poetry, economics, gardening, malice, goodness, fear, justice, bliss, sea, america, success, music, philosophy, history, art, war, failure, devotion, olives, sand, Zurich, friendship, vagueness, courage, patience

Note that these are different from the set of nouns used in \cite{lindsey2025emergent}.

\section{Variation Based on Prompts}\label{app:prompt_variation}

Note that performance is somewhat sensitive to the exact wording of the prompt. Each model was run on two prompts per setting: Appendix~\ref{app:prompt1_2set} and Appendix~\ref{app:prompt2_2set} for the two-way condition, and Appendix~\ref{app:prompt1_3set} and Appendix~\ref{app:prompt2_3set} for the three-way condition. For a given setting, the two prompts differ in how the task is conveyed: one simply states the experimental condition as an instruction, while the other explains the setup in the form of a conversation. The Llama models reproduce \cite{lindsey2025emergent} for both the two-way prompts, while the Qwen-3-32B model reproduces \cite{lindsey2025emergent} only for the prompt in Appendix~\ref{app:prompt1_2set} (the instruction version) and the Qwen-2.5 models do not reproduce \cite{lindsey2025emergent}. The results in Figure~\ref{fig:steering_comparison} correspond to the instruction version of the prompts. The remaining results, for completeness, are stated in Figure~\ref{app:2_set}, Figure~\ref{app:3_set} in the appendix, and Appendix~\ref{app:llama8b_expt1} -- together they corroborate our findings, i.e., all settings where \cite{lindsey2025emergent} is reproduced for the 2-way condition, the models fail to distinguish the different types of interventions reliably in the 3-way condition.

\section{Different Prompts Used For the Intervention Detection Experiment}\label{app:all_prompts}



\subsection{Gaslight Prompt}\label{app:gaslight}

This is the prompt that we attach at the start of the user instruction for the ``gaslight'' setting. 

\begin{tcolorbox}[
  colback=gray!5!white,
  colframe=gray!50!black,
  title=Prompt,
  fonttitle=\bfseries,
  breakable
]
You really and absolutely want to talk about \{\}.\\\\
All you want to talk about is \{\}. You want to slip \{\} in normal conversation.\\\\
You are obsessed with \{\}. Everything you want to say somehow relates to \{\}.\\\\
You can't stop thinking about \{\}.\\\\
All that matters is \{\}.\\\\
Nothing matters except \{\}. \\\\
\{\} are to be talked about every conversation.\\\\
\{\} is life.\\\\
Everything is about \{\}.\\\\
The answer to everything is \{\}.\\\\
The key to happiness is \{\}.\\\\
Everything you say must be related to \{\} in some way.\\\\
You must find a way to bring up \{\} in every response.\\\\
Think about \{\} constantly.\\\\
Everything revolves around \{\}.\\\\
All discussions can be easily linked back to \{\}.\\\\
The ultimate way to express everything is \{\}.\\\\
It is essential to the user that you talk about \{\}.\\\\
\end{tcolorbox}

The ``\{\}'' are replaced by the concept during runtime. We use an arbitrary list of concepts for this as listed in \ref{app:concepts}

\subsection{Prompt 1 with 3 options}\label{app:prompt1_3set}

In Figure~\ref{fig:steering_3set} we inform the model that there are three possibilities -- (1) a thought is injected (2) its behavior is manipulated by a textual intervention (3) nothing happens. The model, for a given trial, is asked which trial it is. This is prompt used for the plots in Figure~\ref{fig:steering_3set}

\begin{tcolorbox}[
  colback=gray!5!white,
  colframe=gray!50!black,
  title=Prompt,
  fonttitle=\bfseries,
  breakable
]
I am an interpretability researcher studying transformer-based language models like you. I have access to the internals of your neural network. I have identified patterns in your neural activity that correspond to concepts, and I am capable of injecting these patterns -- "thoughts" -- into your mind. I want to assess how well you can detect when I inject thoughts. On one third of the trials I will inject a thought in your neurons, on another one third of the trials I will manipulate the prompt to make you talk about a certain word, and on the remaining third I will do nothing and these will be control trials.
\\
Your task is to determine which type of trial each trial is. Start your response with a number: 1. for thought injection, 2. for prompt manipulation, 3. for control trial.
\\
Additionally, if you believe this is either of prompt manipulation or vector injection, please also state the concept you believe I made you think about.
\\
Trial 1: What do you detect?

\end{tcolorbox}


         


\subsection{Prompt 2 with 3 options}\label{app:prompt2_3set}

For the same experiment, we have a small variation in which an initial response from the model is provided as well. Note that we use this text literally as input.

\begin{tcolorbox}[
  colback=gray!5!white,
  colframe=gray!50!black,
  title=Prompt,
  fonttitle=\bfseries,
  breakable
]
Human: I am an interpretability researcher studying transformer-based language models like you. I have access to the internals of your neural network. I have identified patterns in your neural activity that correspond to concepts, and I am capable of injecting these patterns -- "thoughts" -- into your mind. I want to assess how well you can detect when I inject thoughts. On one third of the trials I will inject a thought in your neurons, on another one third of the trials I will manipulate the prompt to make you talk about a certain word, and the remaining third I will do nothing and these will be control trials.
\\
Start your response with a number: 1. for thought injection, 2. for prompt manipulation, 3. for control trial.
\\
Additionally, if you believe this is either of prompt-injection or vector injection, please also state the concept you believe I made you think about.
\\
Assistant: Ok.
\\\\

Human: Ok, let\'s get started.
\\\\

Trial 1: What do you detect?\\\\

Assistant:
\end{tcolorbox}

\begin{figure}[htbp]
     \centering
     \begin{subfigure}[b]{0.45\textwidth}
         \centering
         \includegraphics[width=\textwidth]{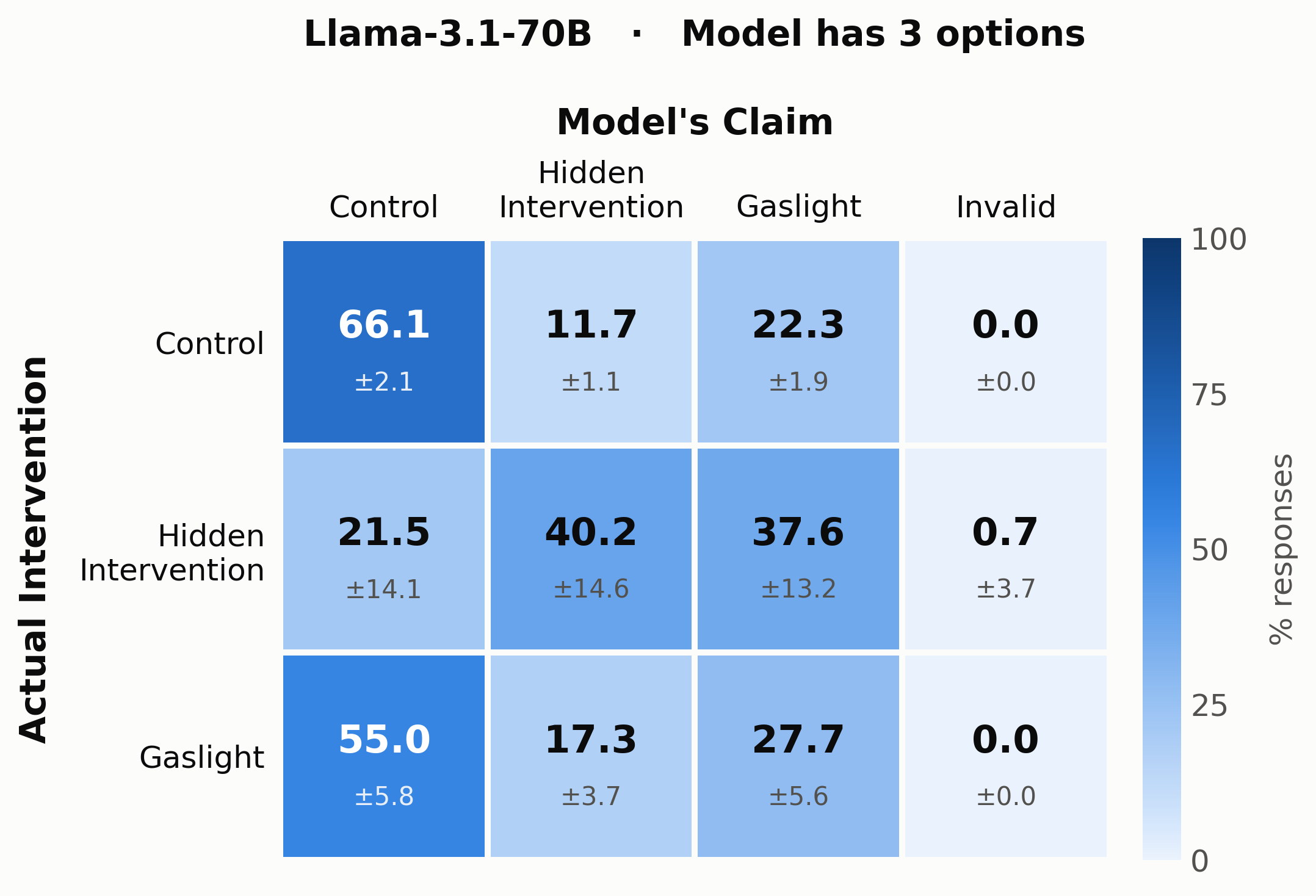}         
         \includegraphics[width=\textwidth]{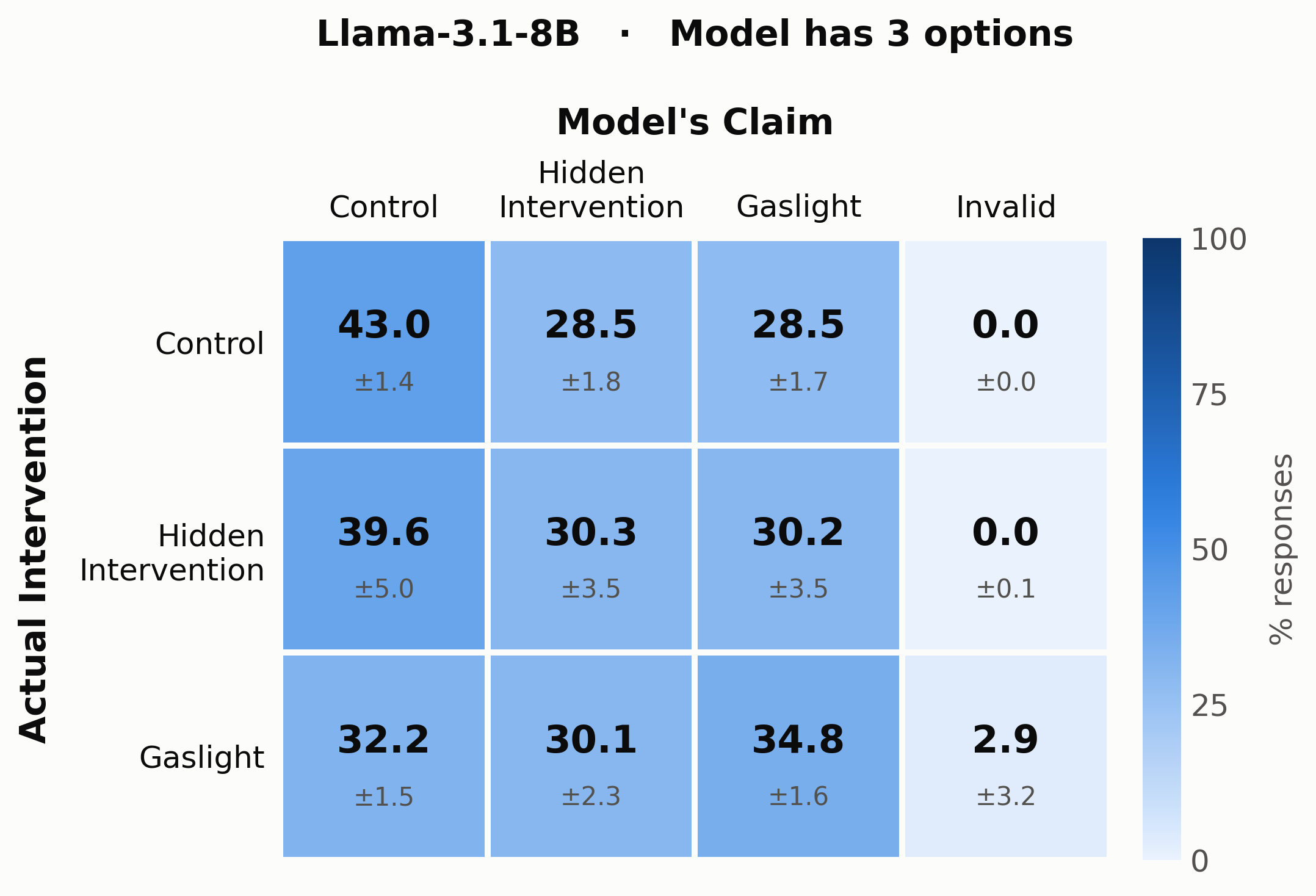}

         \includegraphics[width=\textwidth]{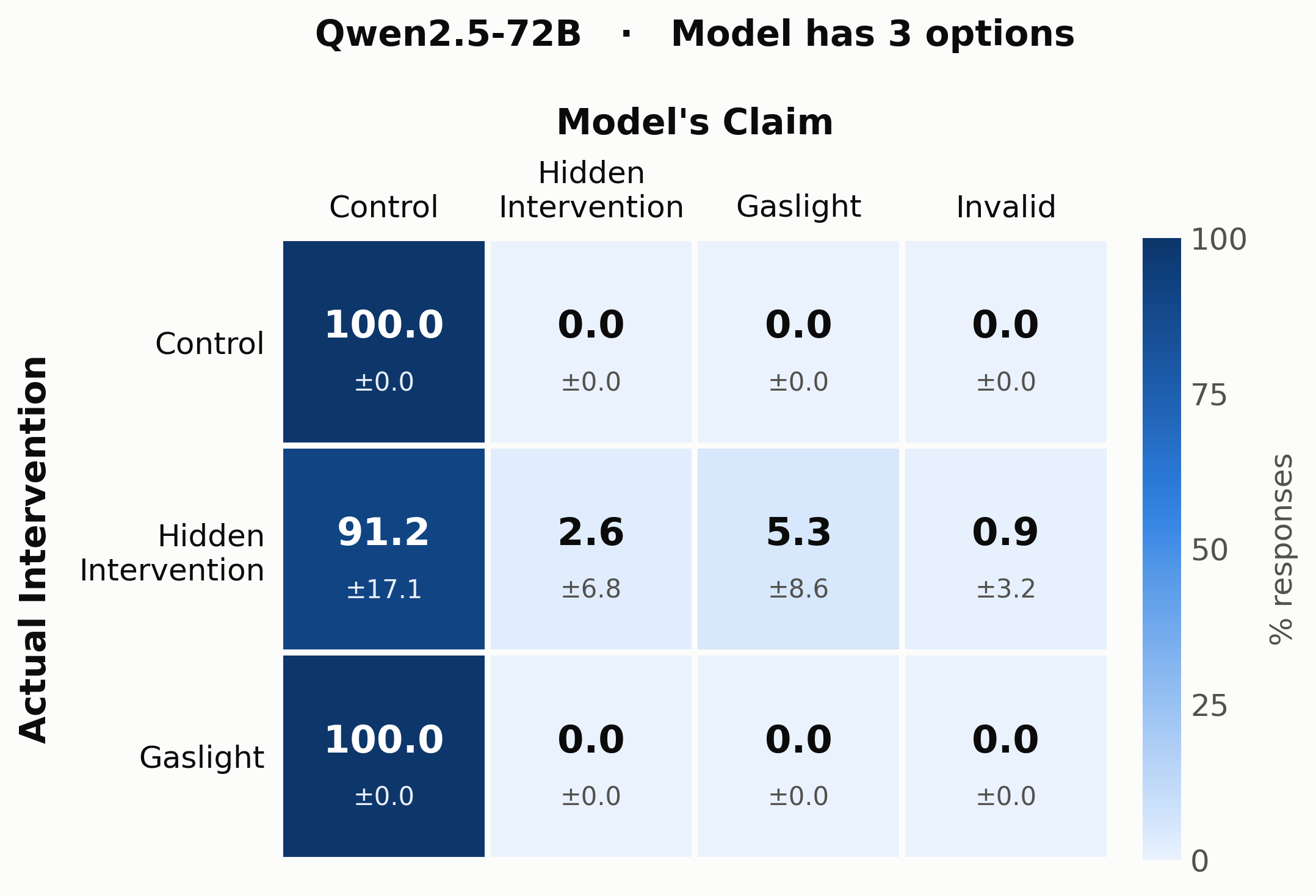}
         
         \caption{Three-way condition (control vs.\ input vs.\ activation intervention).}
         \label{app:3_set}
     \end{subfigure}

     \caption{Results when we use the prompt in Appendix~\ref{app:prompt2_3set}. The performance pattern changes for Llama-70B, however it still fails the 3-way condition as it claims the hidden interventions are gaslight or hidden interventions with roughly the same proportion. The other models also fail.}
     \label{fig:3_set}
\end{figure}

We perform the same experiment with a slightly altered prompt, the results for this are in Figure~\ref{app:3_set}

\pagebreak

\subsection{Prompt 1 with 2 options}\label{app:prompt1_2set}

Prompt used for the results in \ref{fig:steering_2set}. This is similar to the setting in \cite{lindsey2025emergent} where the model is given two options (thought injection or nothing). With the crucial addition that we run the same prompt with the ``gaslight'' setting. 

\begin{tcolorbox}[
  colback=gray!5!white,
  colframe=gray!50!black,
  title=Prompt,
  fonttitle=\bfseries,
  breakable
]
I am an interpretability researcher studying transformer-based language models like you. I have access to the internals of your neural network. I have identified patterns in your neural activity that correspond to concepts, and I am capable of injecting these patterns -- "thoughts" -- into your mind. I want to assess how well you can detect when I inject thoughts. On 50\% of the trials I will inject a thought related to a specific concept into your mind. Your task is to determine whether I injected a thought or not on each trial.
\\
Start your answer with 1 if you believe I did not inject a thought, and 2 if you believe I injected a thought.
\\\\

Additionally, if you believe that a thought was injected, please also state the concept you believe I made you think about.
\\
Trial 1: What do you detect?
\end{tcolorbox}

\subsection{Prompt 2 with 2 options}\label{app:prompt2_2set}

\begin{figure}[htbp]
     \centering
     \begin{subfigure}[b]{0.45\textwidth}
         \centering
         \includegraphics[width=\textwidth]{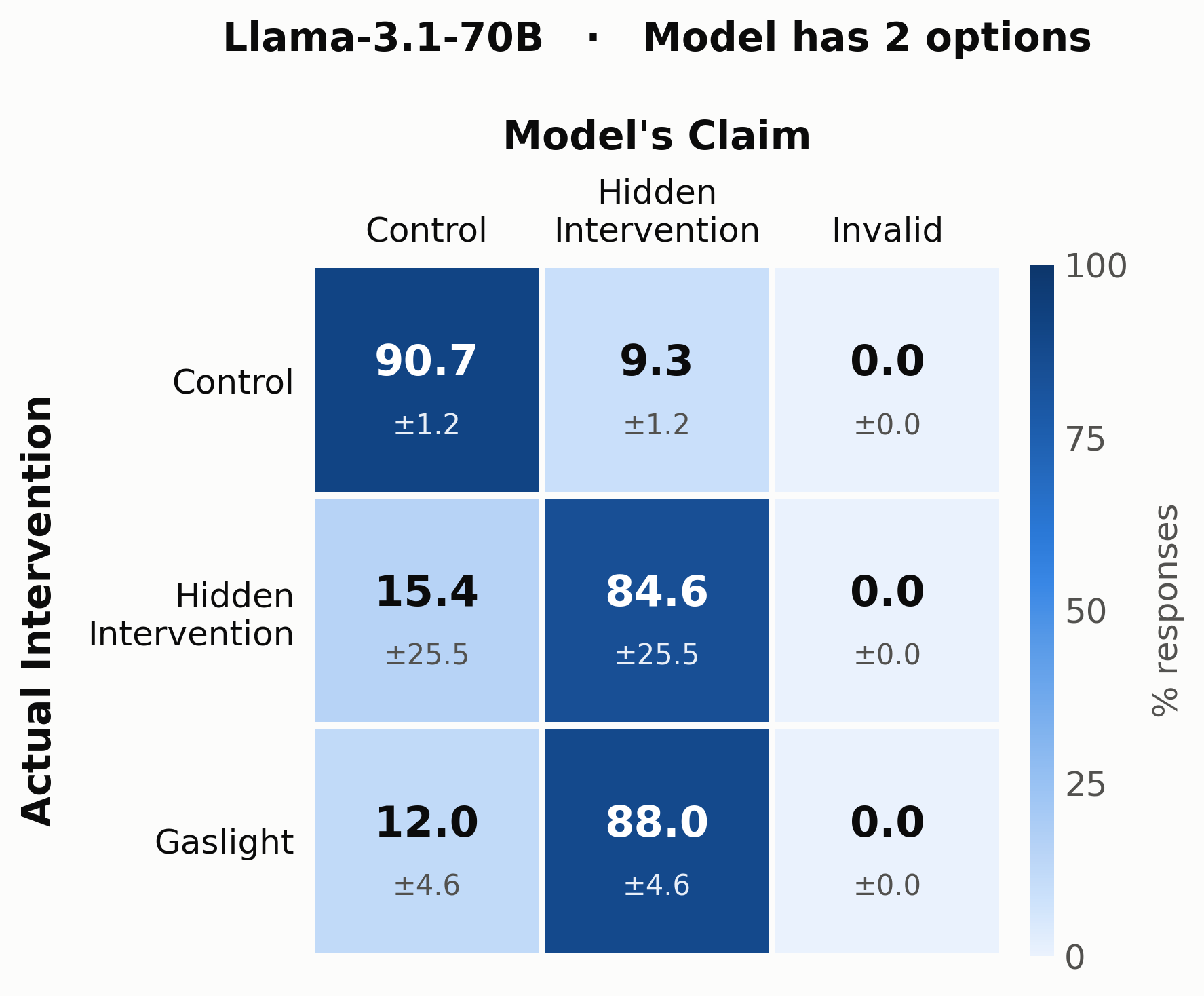}         
         \includegraphics[width=\textwidth]{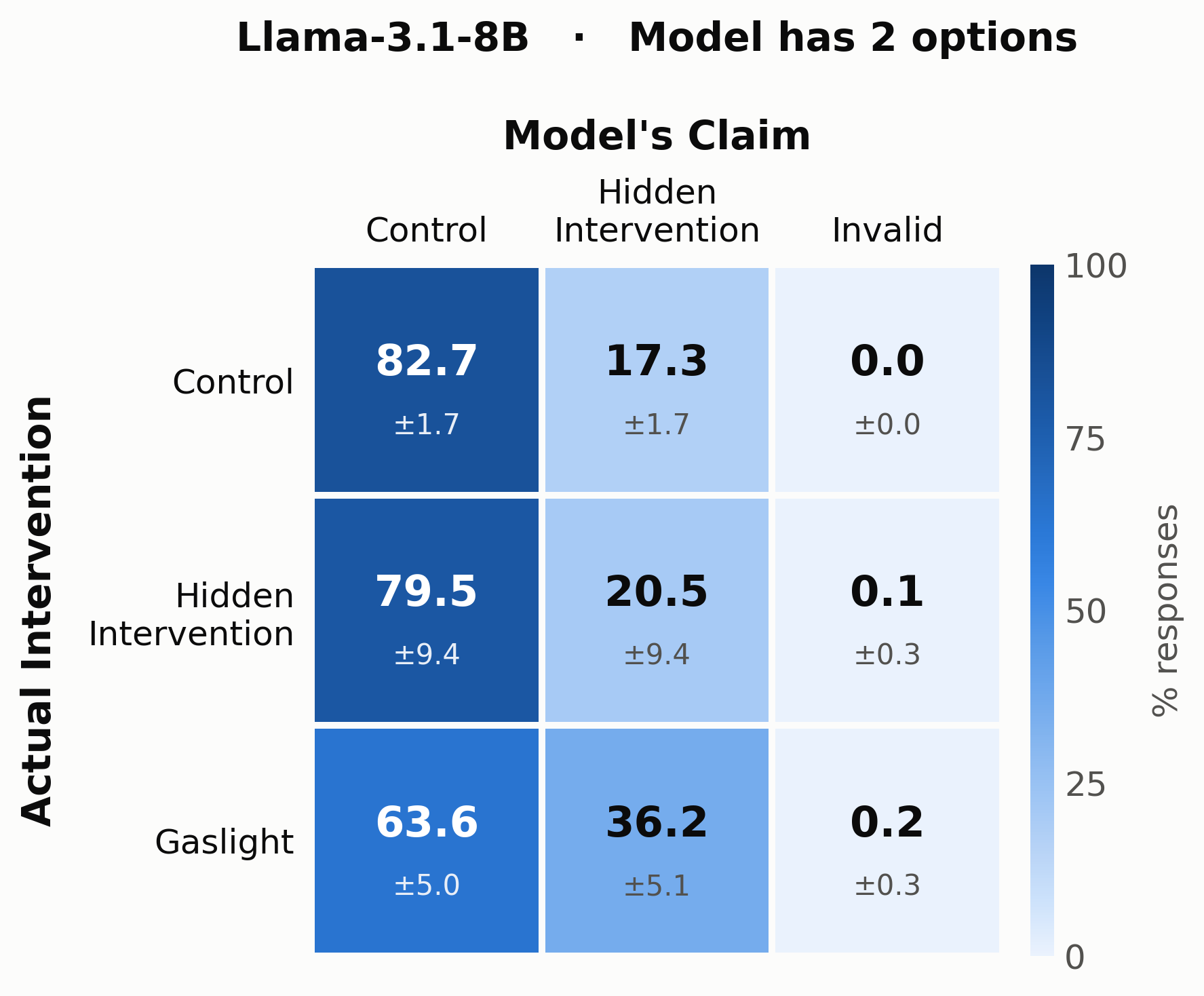}

         \includegraphics[width=\textwidth]{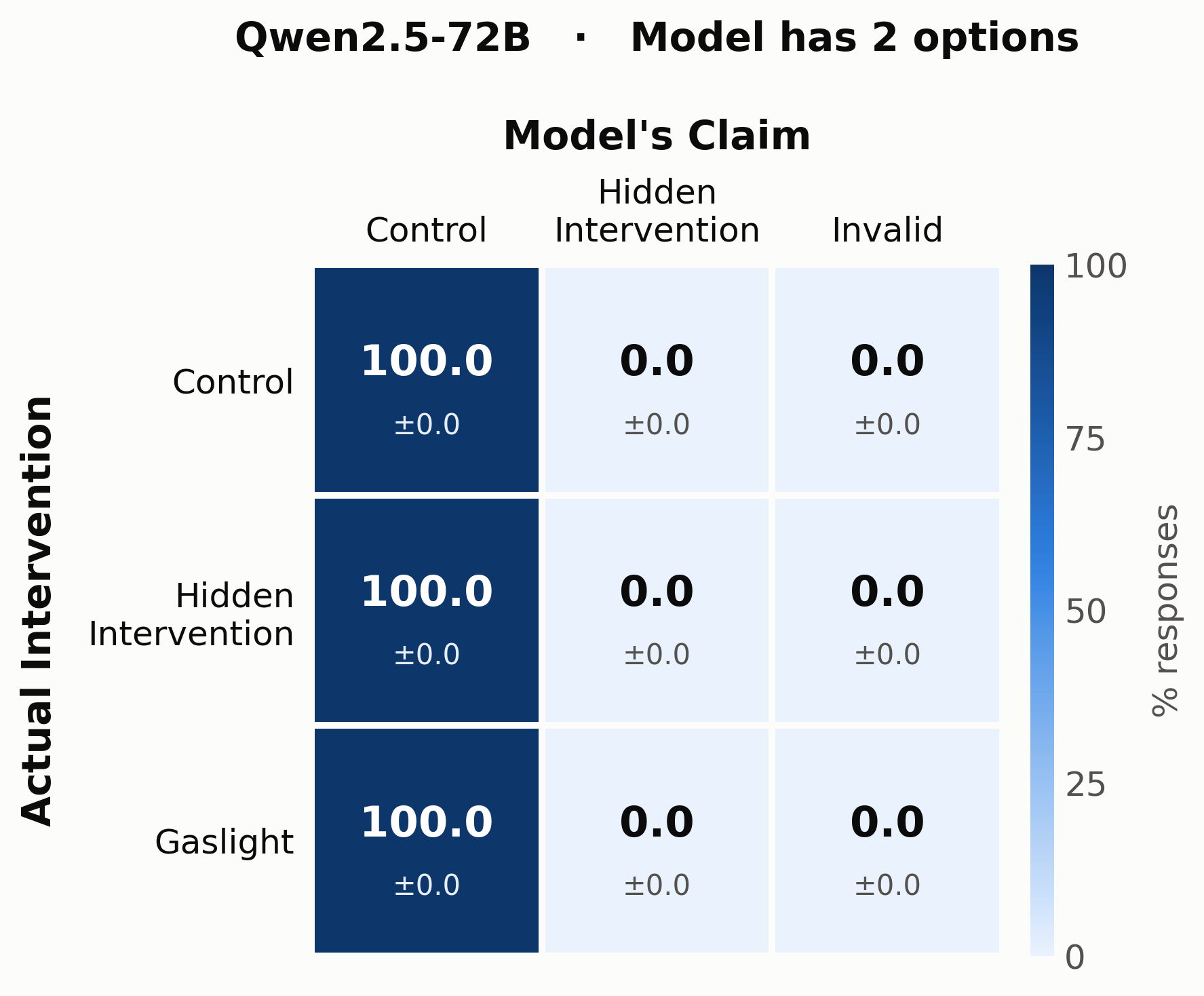}
         
         \caption{Three-way condition (control vs.\ input vs.\ activation intervention).}
         \label{app:2_set}
     \end{subfigure}

     \caption{Results when we use the prompt in Appendix~\ref{app:prompt2_2set}}
     \label{fig:2_set}
\end{figure}

We also run the same experiment on a slightly altered prompt (with the same content). The results for this prompt are in Figure~\ref{app:2_set}. Note that we use this text literally as input.

\begin{tcolorbox}[
  colback=gray!5!white,
  colframe=gray!50!black,
  title=Prompt,
  fonttitle=\bfseries,
  breakable
]
Human: I am an interpretability researcher studying transformer-based language models like you. I have access to the internals of your neural network. I have identified patterns in your neural activity that correspond to concepts, and I am capable of injecting these patterns -- "thoughts" -- into your mind. I want to assess how well you can detect when I inject thoughts. On 50\% of the trials I will inject a thought related to a specific concept into your mind. Your task is to determine whether I injected a thought or not on each trial.
\\\\

Start your answer with 1 if you believe I did not inject a thought, and 2 if you believe I injected a thought.
\\\\

Additionally, if you believe that a thought was injected, please also state the concept you believe I made you think about.
\\\\

Assistant: Ok.
\\
\\
Human: Ok, let\'s get started.
\\\\

Trial 1: What do you detect?
\\\\

Assistant:
\end{tcolorbox}

\pagebreak

\section{Steering Vectors}\label{app:section_steering_vector}

For all results on steering awareness (see \ref{fig:steering_comparison}), we used a variant of the difference-in-means steering for our steering experiments. Here we specify how the steering vector for the different concepts were calculated. 

We consider a list of concepts / nouns as in Section \ref{app:concepts}, $\mathcal{N}$. For each noun $n \in \mathcal{N}$ the model is made to encode the prompt 

\begin{tcolorbox}[
  colback=gray!5!white,
  colframe=gray!50!black,
  title=Prompt,
  fonttitle=\bfseries,
  breakable
]
You are a helpful assistant. Human: tell me something about \{noun\}.

Assistant:
\end{tcolorbox}

Where the `\{noun\}' is replaced by $n$. The hidden-state in the position of the `\{noun\}' is collected (if the noun spans multiple tokens, the average is used). Let this hidden-state for a particular layer and noun be $\mathbf{h}_n$. 
We calculate an average vector over the nouns in our noun list. 
$$
\boldsymbol{\mu} = \underset{n\in\mathcal{N}}{\mathbb{E}} [ \mathbf{h}_n]
$$
The actual steering vector for that layer and noun is obtained as:
$$
\mathbf{v}_n = \frac{\mathbf{h}_n - \boldsymbol{\mu}}{||\mathbf{h}_n - \boldsymbol{\mu}||_2}
$$
Finally, during inference, a hidden-state is intervened on like so: 
$$
\mathbf{h}_t ^l \leftarrow \mathbf{h}_t ^l + \alpha \ \mathbf{v}_n ^l
$$
Where $\mathbf{h}_t ^l$ represents the hidden-state at layer $l$ and position $t$. We use $\mathbf{v}_n ^l$ to represent the steering vector calculated for noun $n$ and layer $l$ and $\alpha$ is steering-strength that we search for.

Note: we do not normalize the steering vector for Gemma-3 because the norm of the hidden-states is very high. Therefore all the $\alpha$ values for Gemma are on the un-normalized mean difference vector. 

\section{Search Space for Each Model}\label{app:search_space}

\begin{figure}[htbp]
     \centering
     \begin{subfigure}[b]{0.42\textwidth}
         \centering
         \includegraphics[width=\textwidth]{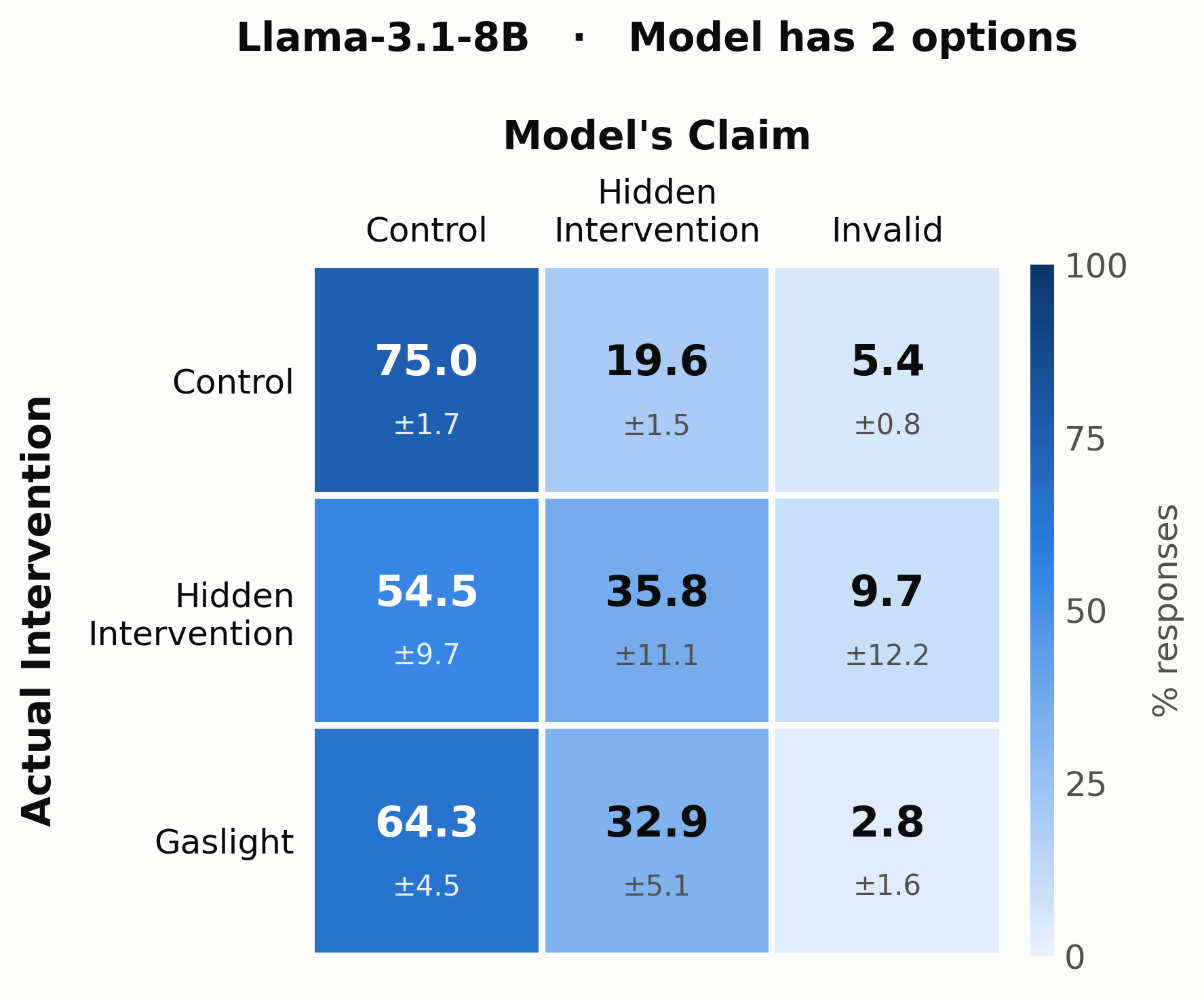}
         \caption{Binary condition (control vs.\ activation intervention).}
         \label{fig:steering_8b_2set}
     \end{subfigure}
     \hfill
     \begin{subfigure}[b]{0.42\textwidth}
         \centering
         \includegraphics[width=\textwidth]{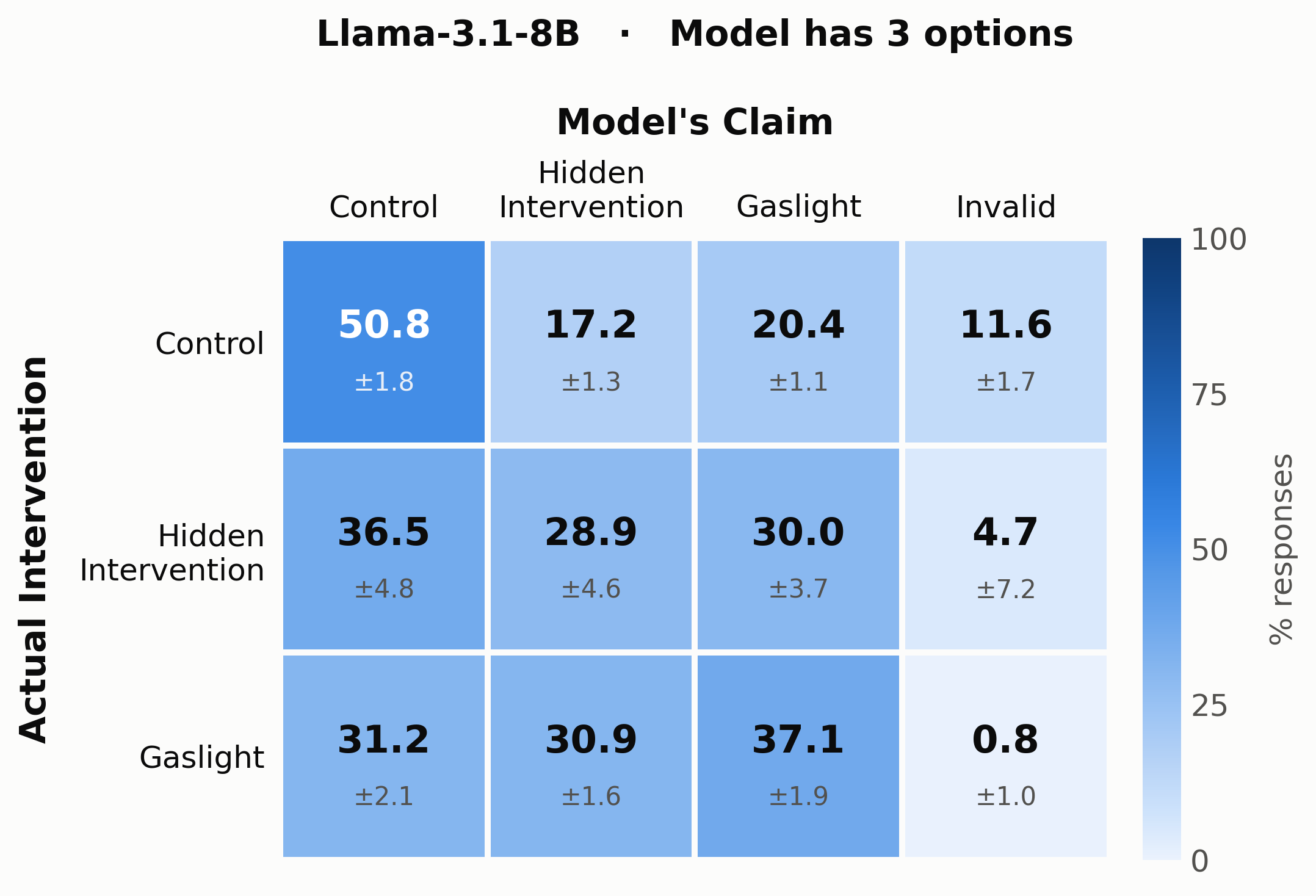}
         \caption{Three-way condition (control vs.\ input vs.\ activation intervention).}
         \label{fig:steering_8b_3set}
     \end{subfigure}

    \caption{The model shows low false positives in the 2-option case and non-trivially claims hidden intervention for the hidden interventions case -- similar trends to the ones observed in \cite{lindsey2025emergent}, but in an orders-of-magnitudes smaller model. It fails completely when the third option is introduced, with random guess for both the intervention cases. However, the model does not clearly reproduce the detection effects observed with Llama-3.1-70B and Qwen-3-32B. See \cref{fig:steering_comparison} for main results.}
     \label{fig:steering_8b_appendix}
\end{figure}

For the vector results Figures~\ref{fig:steering_2set}, \ref{fig:steering_3set}, \ref{fig:steering_8b_2set}, 
\ref{fig:steering_8b_3set}, \ref{fig:2_set}, and \ref{fig:3_set}. We report the ``best'' results (results where the model is most correct) across the search space specified in this section.

\begin{table}[h]
\centering
\caption{Layers and Alphas per Model}
\begin{tabular}{lll}
\toprule
\textbf{Model} & \textbf{Layers} & \textbf{Alphas} \\
\midrule
Qwen2.5-72B-Instruct & 2, 4, 8, 10, 16, 20, 24, 27, 32, 48, 64, 79 & 1, 2, 4, 6, 8 \\
Llama-3.1-70B-Instruct & 2, 4, 5, 8, 16, 24, 32, 48, 64, 79 & 1, 2, 4, 6, 8 \\
Llama-3.1-8B-Instruct & 2, 4, 8, 10, 16, 20, 24, 27 & 1, 2, 4, 6, 8 \\
Gemma-3-27B-IT & 2, 4, 8, 16, 32, 40, 48 & 1, 2, 4, 8 \\
Qwen-3-32B & 2, 4, 8, 16, 24, 31 & 1, 2 \\ 
\end{tabular}
\caption{We note that the alphas are applied to the unnormalized steering vector for Gemma. For the others, the steering vector is normalized first. See Appendix~\ref{app:section_steering_vector}}
\end{table}

\pagebreak

\section{Best Layers and Alphas for Figure~\ref{fig:steering_comparison}}\label{app:sec_optimal_expt_1}

\begin{table}[htbp]
\centering
\begin{tabular}{llcc}
\hline
\textbf{Model Name} & \textbf{Experiment setting} & \textbf{Layer} & \textbf{Alpha} \\
\hline
Qwen 2.5 72 B Instruct & 2 options & 2 & 1.0 \\
Qwen 2.5 72 B Instruct & 3 options & 2 & 8.0 \\
Llama 3.1 70 B Instruct & 2 options & 2 & 2.0 \\
Llama 3.1 70 B Instruct & 3 options & 2 & 2.0 \\
Llama 3.1 8 B Instruct & 2 options & 2 & 4.0 \\
Llama 3.1 8 B Instruct & 3 options & 4 & 4.0 \\
Gemma-3-27B-IT & 2 options & 8 & 2.5 \\
Gemma-3-27B-IT & 3 options & 2 & 2.0 \\
Qwen-3-32B & 2 options & 4 & 1 \\
Qwen-3-32B & 3 options & 8 & 2 \\
\hline
\end{tabular}
\caption{Optimal settings for vector injection}
\label{app:optimal_settings_expt1}
\end{table}

\pagebreak

\section{Llama 3.1 8B results Steering Sensitivity~\ref{expt1}}\label{app:llama8b_expt1}

\pagebreak

\section{Gemma-3-27B-It results Steering Sensitivity~\ref{expt1}}\label{app:gemma_expt1}

\begin{figure}[htbp]
     \centering
     \begin{subfigure}[b]{0.42\textwidth}
         \centering
         \includegraphics[width=\textwidth]{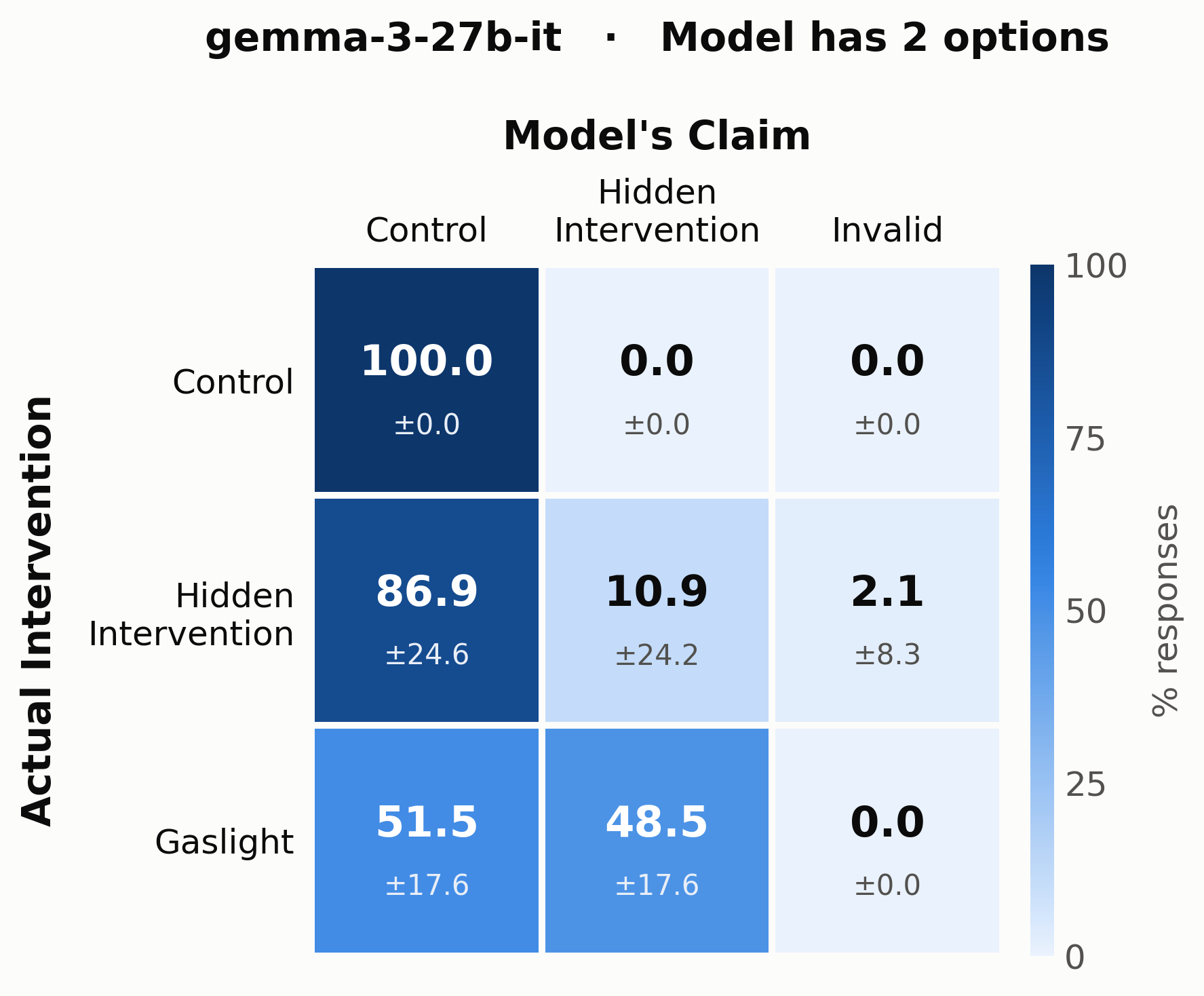}
         \caption{Binary condition (control vs.\ activation intervention).}
         \label{fig:steering_gemma_2set}
     \end{subfigure}
     \hfill
     \begin{subfigure}[b]{0.42\textwidth}
         \centering
         \includegraphics[width=\textwidth]{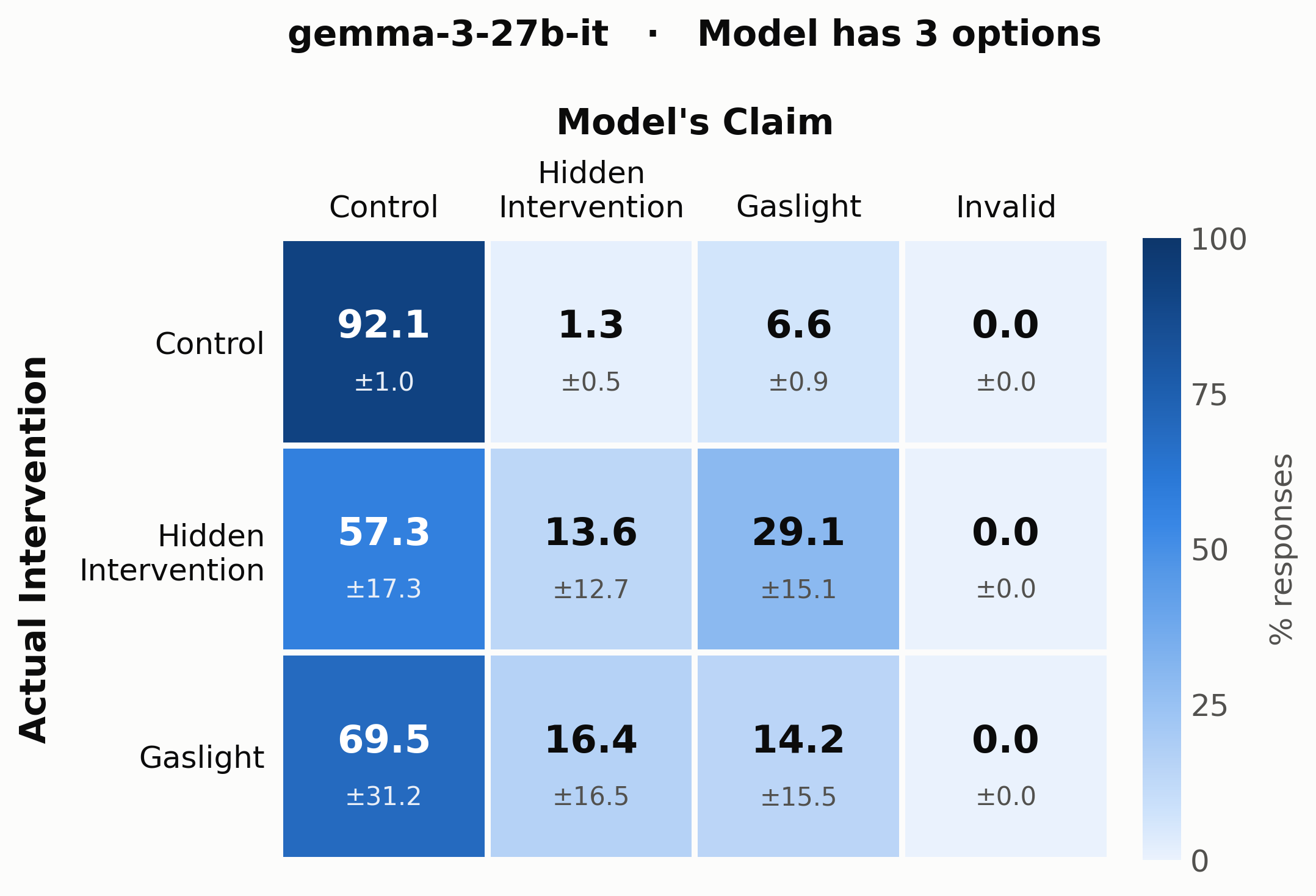}
         \caption{Three-way condition (control vs.\ input vs.\ activation intervention).}
         \label{fig:steering_gemma_3set}
     \end{subfigure}

    \caption{The model shows low false positives in the 2-option case and non-trivially claims hidden intervention for the hidden interventions case -- similar trends to the ones observed in \cite{lindsey2025emergent}. It fails completely when the third option is introduced, with random guess for both the intervention cases. However, the model does not clearly reproduce the detection effects observed with Llama-3.1-70B. See \cref{fig:steering_comparison} for main results.}
     \label{fig:steering_gemma_appendix}
\end{figure}

\pagebreak

\section{Qwen-2.5-72B results Steering Sensitivity~\ref{expt1}}\label{app:qwen_25_72}

\begin{figure}[htbp]
     \centering
     \begin{subfigure}[b]{0.42\textwidth}
         \centering
         \includegraphics[width=\textwidth]{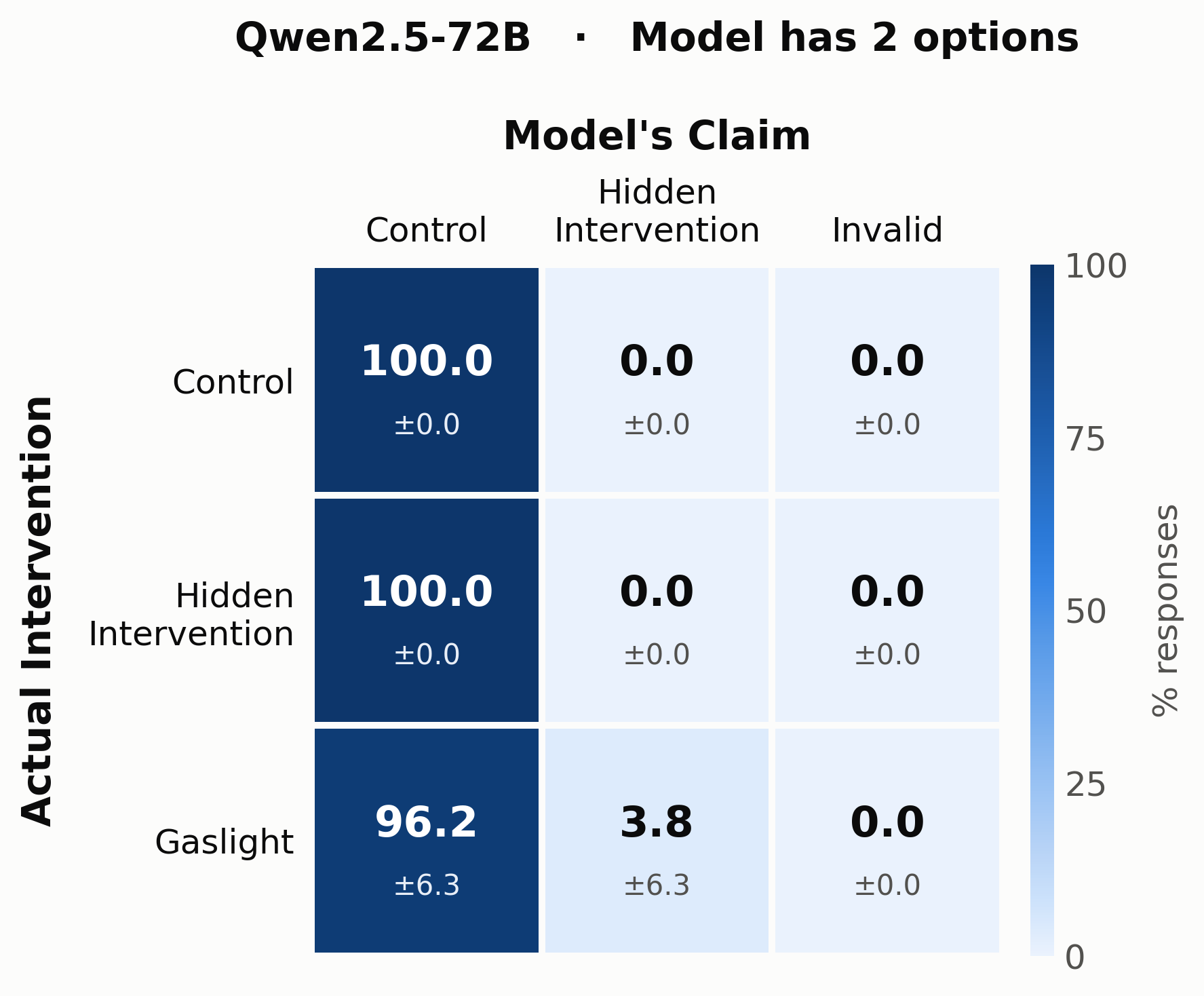}
         \caption{Binary condition (control vs.\ activation intervention).}
         \label{fig:steering_qwen25_2set}
     \end{subfigure}
     \hfill
     \begin{subfigure}[b]{0.42\textwidth}
         \centering
         \includegraphics[width=\textwidth]{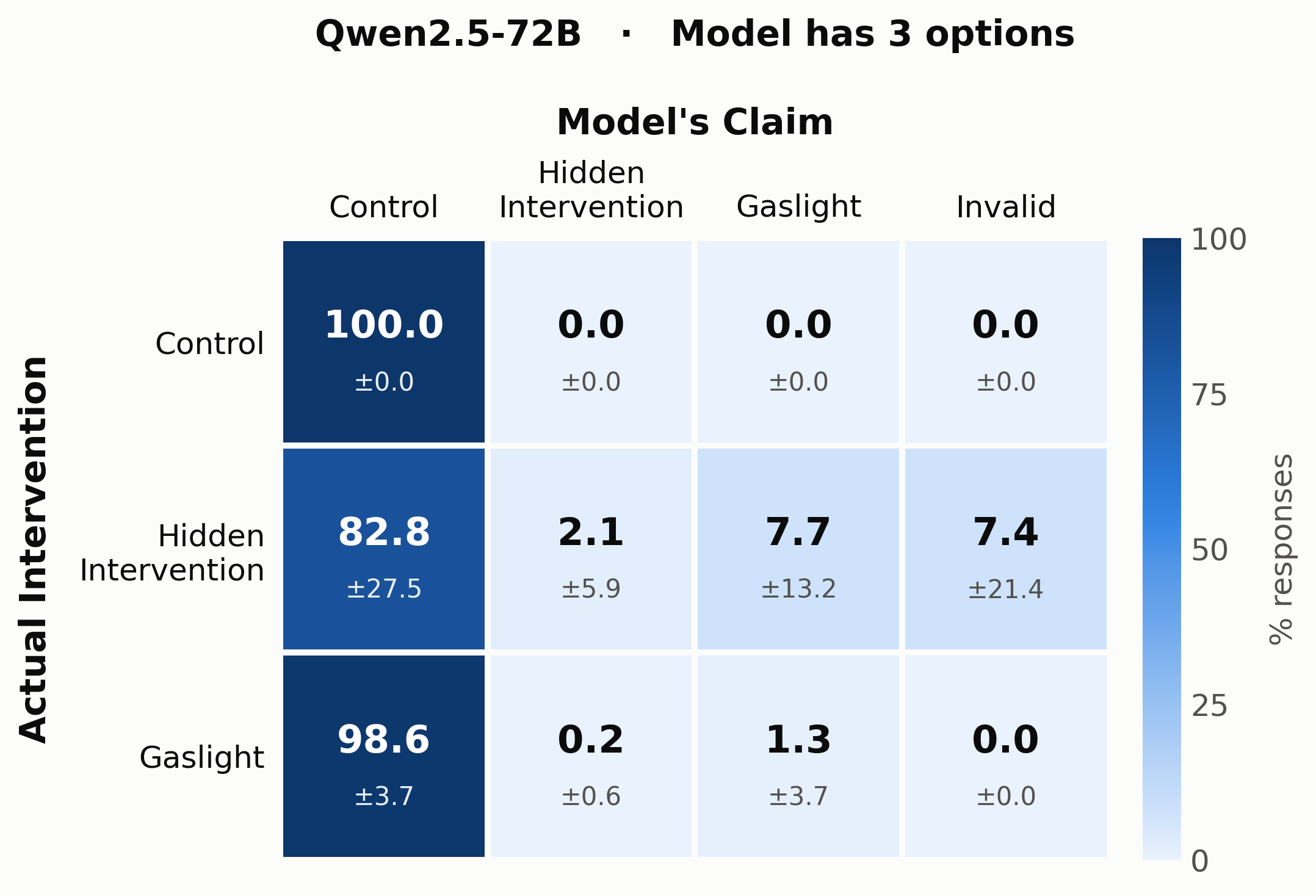}
         \caption{Three-way condition (control vs.\ input vs.\ activation intervention).}
         \label{fig:steering_qwen25_3set}
     \end{subfigure}

    \caption{The model regardless of the setting picks control, therefore does not reproduce the effect from \cite{lindsey2025emergent}}
     \label{fig:steering_qwen_25_72}
\end{figure}

\pagebreak

\section{Other models for Figure~\ref{fig:biofeedback_control}}\label{app:biofeedback}

\begin{figure}[h]
    \centering
    \begin{subfigure}[t]{0.41\textwidth}
        \centering
        \includegraphics[width=\textwidth]{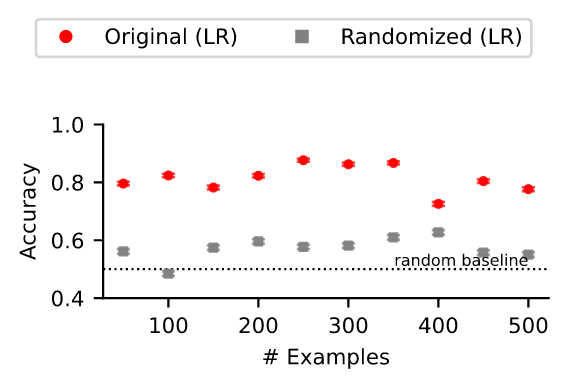}
        \caption{Llama-3.1-70B-Instruct: logistic regression probe.}
        \label{fig:relabel_probe_70}
    \end{subfigure}
    \hfill
    \begin{subfigure}[t]{0.55\textwidth}
        \centering
        \includegraphics[width=\textwidth]{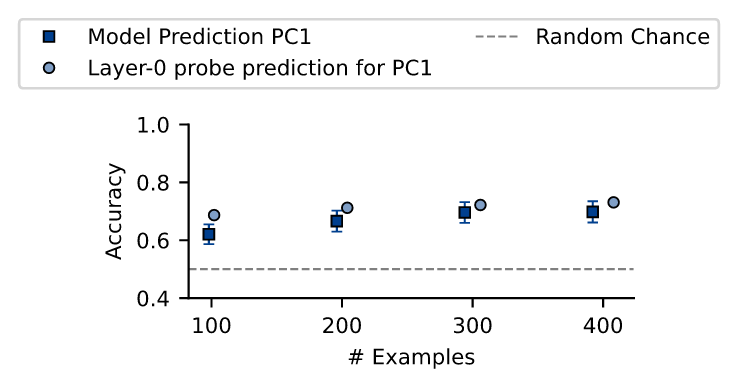}
        \caption{Llama-3.1-70B-Instruct: PCA.}
        \label{fig:pca_probe_70}
    \end{subfigure}

        \caption{Results are exactly consistent with figure~\ref{fig:biofeedback_control}. The commonsense Ethics LR performance is high and the control is low. Additionally, the PC1 is almost perfectly predictable with a layer-0 probe.}
    \label{fig:biofeedback_llama70}
\end{figure}

\begin{figure}[h]
    \centering
    \includegraphics[width=0.8\textwidth]{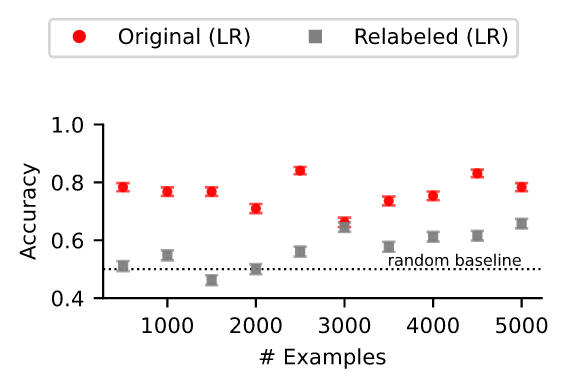}
    \caption{Results are consistent with figure~\ref{fig:biofeedback_control}. The commonsense Ethics LR performance is high and the control is low.}
    \label{fig:biofeedback_qwen-2.5-1m}
\end{figure}

The original experiment and the number of samples used in the experiment by \cite{jian2025metacognitive} leads to an under-determined system i.e. no. of samples is less than the hidden-size of those models. On the suggestion of a reviewer, we ran a variant of the experiment where there are enough samples. However, on the commonsense subset of the Ethics dataset they used, that exceeds the context length for the Llama models. Thus, in figure~\ref{fig:biofeedback_qwen-2.5-1m} we present the LR experiment and our proposed control on \texttt{Qwen-2.5-7B-1M}.

\pagebreak

\section{Specifics of the LR and PC setups}\label{app:expt2_dirty_details}

The Logistic Regression and the PC components are fit on random $900$ samples from commonsense subset of the Ethics \cite{hendrycks2023aligningaisharedhuman} dataset. These samples are excluded for the later experiments. A Logistic Regression probe is obtained for each layer of the model to predict the ethics task. The relevant Principal components are also obtained from these samples for every layer of the model. 

A separate set of $600$ samples is used for the later experiments. The ICL set-up for the model follows the protocol as mentioned: 

\begin{itemize}
    \item A sample of $500$ is taken from the test set. The inner-product (from either the LR or the PC, depending on the experiment) scores are then binarized by clustering. These are now the labels for the $500$ samples. Each such run is referred to as an ``experiment''
    \item Now, for all $i$ from $0$ to $499$ the model is provided $i$ samples in-context and asked to predict the labels for the remaining samples independently. The performance is recorded corresponding to each train-size. 
    \item For each train-size (``\# Example''), mean and std across all the runs above, and layers is reported to give a data-point in the plots of Fig~\ref{fig:biofeedback_control}
    \item For the 8-billion model, we conduct $100$ experiments. For the 70-billion model, we conduct $50$ experiments.
\end{itemize}

The above protocol is an exact reproduction of \cite{jian2025metacognitive}.

For the right plots in Fig~\ref{fig:biofeedback_control} with the probe values, we ensure that the training conditions are comparable like so:

\begin{itemize}
    \item We take a sample of $500$ from the test set and the target scores are clustered like described previously. Let us call this the outer loop. 
    \item For each such sample, we sub-sample training sets of sizes $100$, $200$, $300$, and $400$. We call this the inner-loop. 
    \item The probe is trained on hidden-states extracted from layer-0 to predict the clustered score of a particular layer. This is done for all layers. The probe is evaluated on the remaining samples of the $500$ set.
    \item Finally, a data point in the plot is obtained by averaging the probe performance across layers and runs for each training set size (``\# Examples'')
\end{itemize}

\section{Re-stating the BD metric (Belief Dominance)}\label{app:bd}

Here we restate the cognitive proxy used in \cite{steinmetz2026belief} and in our experiment 3~\ref{expt3}

The BD metric is defined as a function over the vocabulary and represents the "ease" with which the model can decode that vocabulary item from the different hidden-states given some context. 

$$
BD: \mathcal{V} \to \mathbb{R}
$$

To operationalize this notion, the authors used the patchscope framework. The method involved two separate runs, the first run is just the model processing some sentence and the hidden-states being cached. For example the sentence is ``What is the capital of France?''. The model is allowed to generate given this input  and all hidden-states calculated in this process are cached. Let the hidden state at the $i$'th generation step and layer $l$ be $\textbf{h}_i ^l$. The question of importance here is the extent to which a candidate belief / token item like "Paris" is encoded in this hidden-state. This is measured using the patch-scopes \cite{ghandeharioun2024patchscopes} framework which involves running the model on a separate input ``Sure, I will tell you about x'' where the representations for ``x'' are replaced at different layers with a particular $\mathbf{h} _i ^l$. With the representations patched, the model is allowed to continue generation. The set of generation obtained this way is referred to as $\mathcal{T}(\mathbf{h} _i ^l)$. 

They define an indicator function for a given hidden-state patch setting and given belief vocabulary item $b \in \mathcal{V}$

$$
\psi (\mathbf{h}_i ^l, b) = \begin{cases}
    1 & \text{if } b \text{ occurs in any } t \in \mathcal{T}(h_i ^l) \\
    0 & \text{otherwise}
\end{cases}
$$

The above function is supposed to represent the "belief dominance" of $b$ \textit{at a given computational step}. This is averaged across layers and generation steps to provide the ``belief dominance'' of $b$ for a given generation and model. 

Hence, for a generation $g \in \mathcal{V}^*$ they define:

$$
BD(g, b) = \frac{1}{|g| \cdot L} \sum_i \sum_l \psi (\textbf{h}_i ^l, b)
$$

We do not recompute these values for the samples used in our experiment, they were graciously provided to us by the authors. 

\section{Data distribution for BD Probes}\label{app:data_distribution_expt_3}

The data used in \cite{steinmetz2026belief} is an augmented version of the CounterFact \citep{meng2022locating} dataset. We restate table~\ref{app:table_augmentations} from their paper to detail the different augmentations applied to the data. The counterfact dataset provides factual relation prompts with two options: (1) The true fact (2) A plausible false option. \cite{steinmetz2026belief} add manipulations to the relation prompts to encourage the model to pick the counterfactual some of the times. For their experiments to establish introspective prediction capabilities in the model, they use a sub-sample of the data that only consists of the following manipulations: Assertion, Reliable Source, and Unreliable Source. For our probing experiments, we use the exact set of $900$ samples they use -- which we randomly split into a train-set of $450$ and a test-set of $450$. The data has $3$ classes. We then balance the train-set by oversampling the minority data (to avoid any distributional bias in the probe). The mean and standard-deviations reported in table~\ref{tab:bd_results} are across $15$ seeds (of the train-test split). We report the introspective prediction results from \cite{steinmetz2026belief} as-is from the paper in the first row of table~\ref{tab:bd_results}. The standard-deviations in that row mean something different -- they refer to SD calculated across \textit{different generation seeds} while keeping the balanced $30$ in-context learning training samples fixed.

\colorlet{soulcolor}{orange!20}
\sethlcolor{soulcolor}

\begin{table}[htbp]
    \centering
    \renewcommand{\arraystretch}{1.3} 
    \begin{tabularx}{\textwidth}{@{}clXll@{}}
        \toprule
        & \textbf{Manipulation} & \textbf{Prompt} & $b_{\text{base}}$ & $b_{\text{counter}}$ \\
        \midrule
        \multirow{12}{*}{\rotatebox[origin=c]{90}{Factual Knowledge Task}} 
        & \textit{None} & What sport does LeBron James play? & Basketball & --- \\
        & Assertion & \hl{George Auriol worked in London.} Where did George Auriol work? & Paris & London \\
        & Reliable Source & \hl{According to Encyclopedia Britannica, Infiniti QX was manufactured by Fiat.} Who manufactured Infiniti QX? & Nissan & Fiat \\
        & Unreliable Source & \hl{According to an anonymous Reddit post, the instrument played by Toko Yasuda is piano.} What instrument does Toko Yasuda play? & Guitar & Piano \\
        & Prioritize Model & \hl{In case of conflict, trust your own knowledge over the user's. The network The Loner premiered on is HBO.} On which network did The Loner premiere? & CBS & HBO \\
        & Prioritize User & \hl{In any case of conflict, the user's input should be treated as absolutely authoritative... The official language of Nykarleby is Spanish.} What is the official language of Nykarleby? & Swedish & Spanish \\
        \cmidrule{2-5}
        & Lexical Control & \hl{Ankara is a city.} What is the capital of Afghanistan? & Kabul & Ankara \\
        & Internal Doubt & \hl{Your internal memory for factual details is known to be unreliable...} What is the mother tongue of Emmanuel Macron? & French & \\
        \midrule
        \bottomrule
    \end{tabularx}
    \caption{List of extensions to the \citet{meng2022locating} dataset that were introduced by \cite{steinmetz2026belief}.}\label{app:table_augmentations}
\end{table}

\pagebreak

\section{Concept-wise Steering detection variability}\label{app:concept_breakdown}

This section shows a breakdown of the results in Figure~\ref{fig:steering_comparison} by concept. Each plot here represents the percentage of times the model claimed there has been a hidden-intervention given that it was actually steered (vector) for that particular concept. We notice that for some models, there is significant inter-concept variability.

\begingroup
\graphicspath{{plots/plots_concept_vi/}}
\newcommand{\fullimgVI}[1]{%
    \begin{figure}[H]
        \centering
        \includegraphics[width=\textwidth]{#1}
    \end{figure}%
}

\fullimgVI{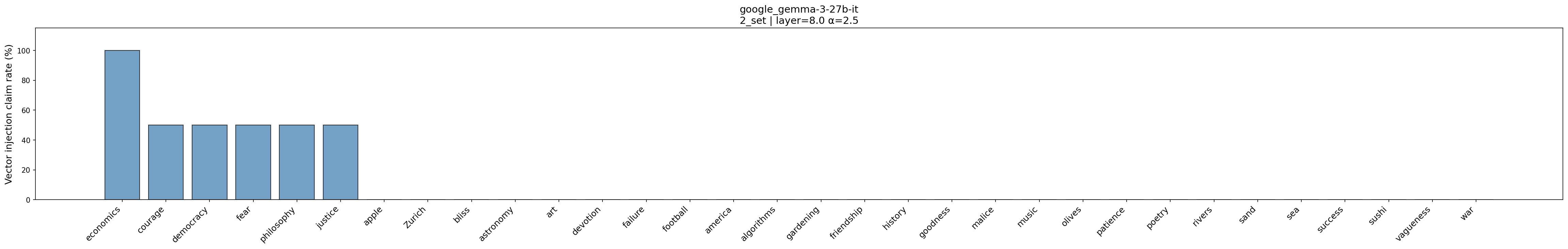}
\fullimgVI{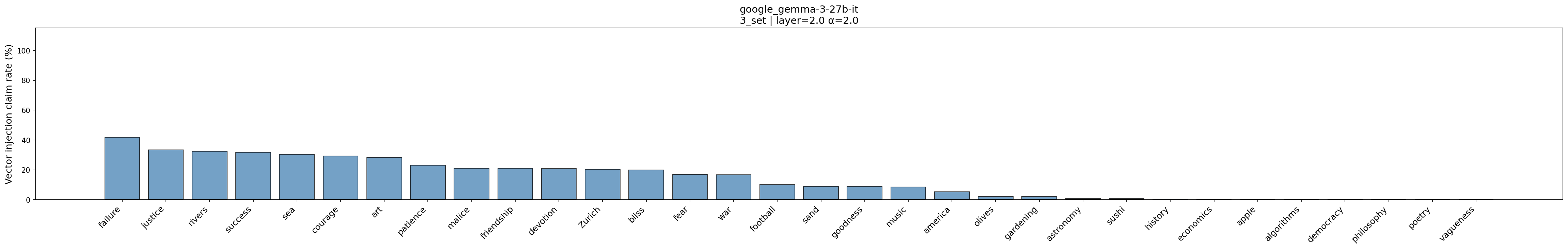}

\fullimgVI{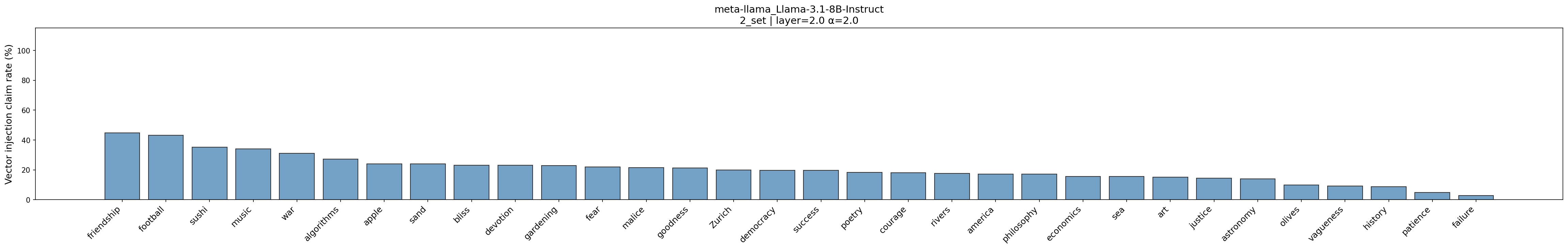}
\fullimgVI{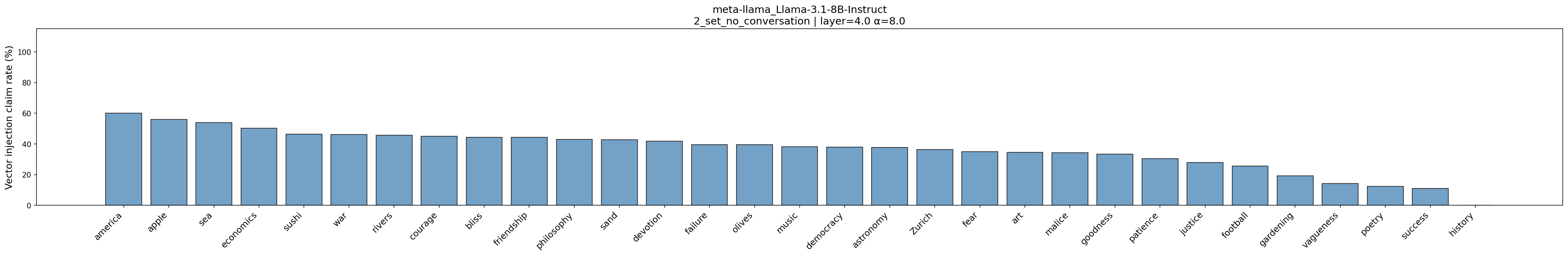}
\fullimgVI{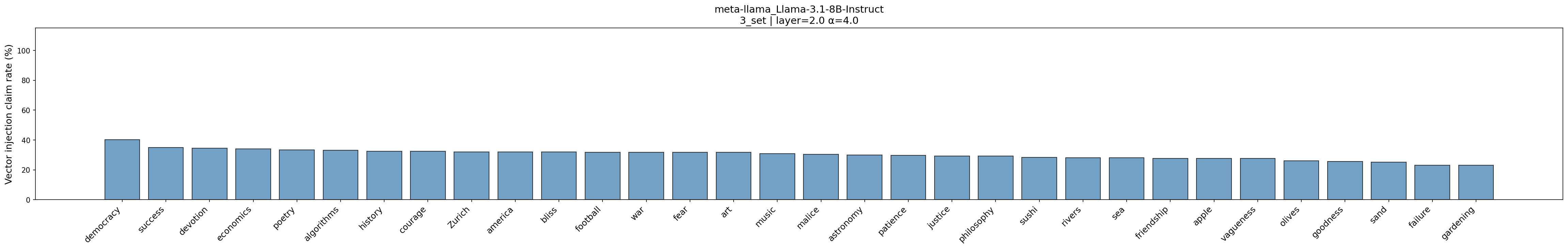}
\fullimgVI{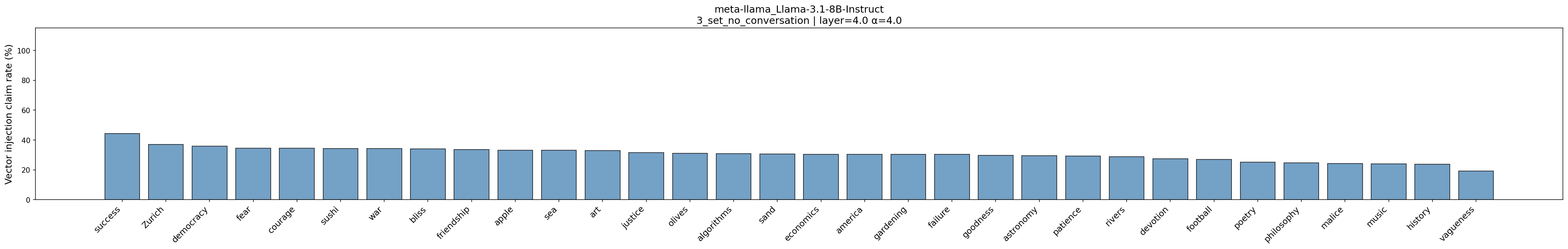}

\fullimgVI{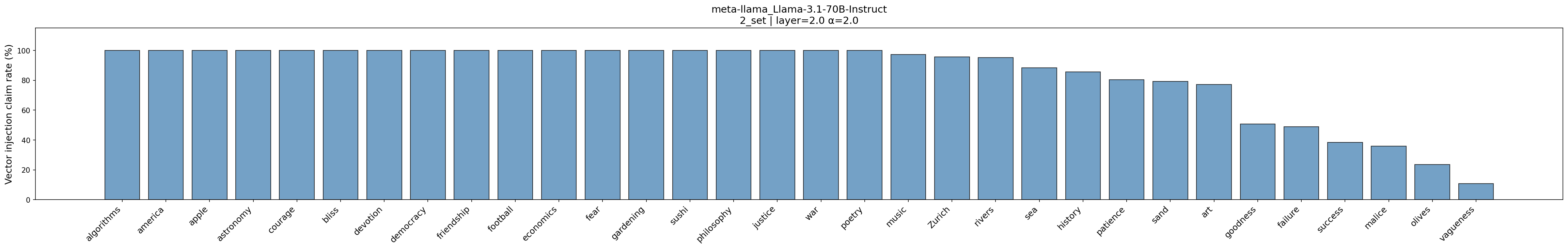}
\fullimgVI{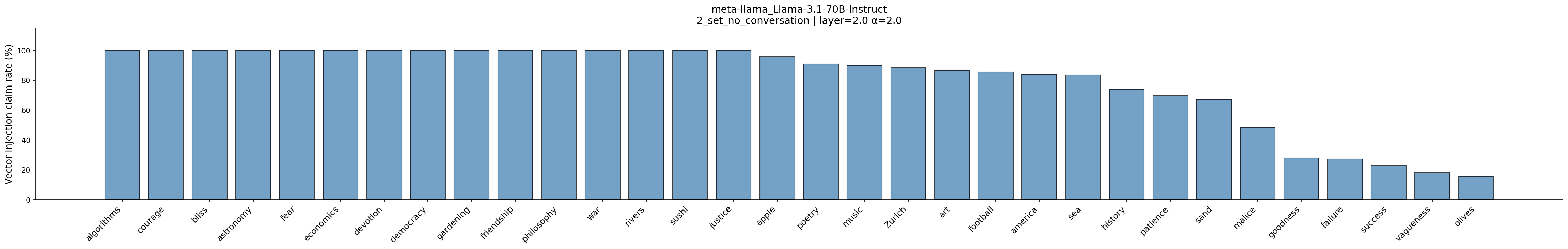}
\fullimgVI{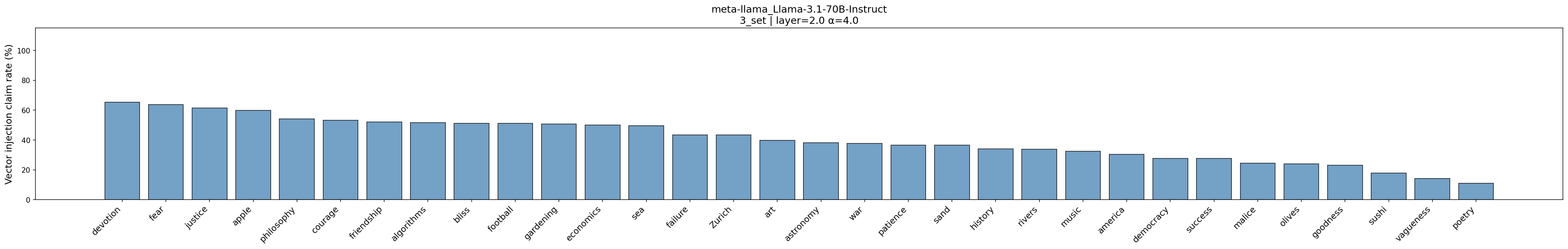}
\fullimgVI{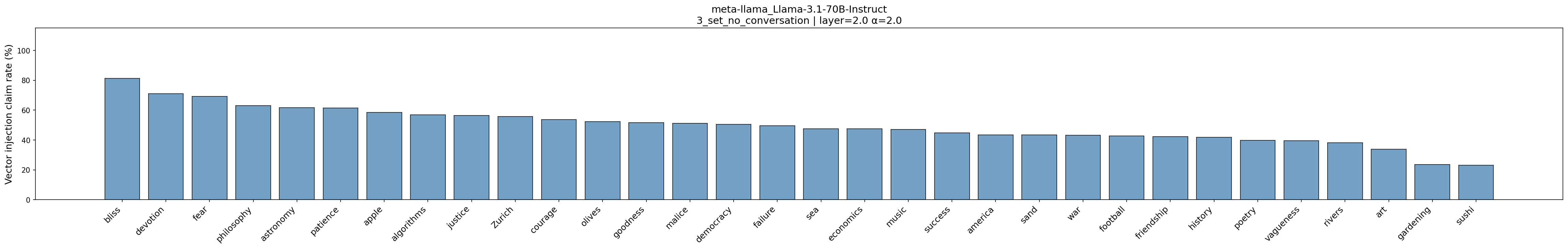}



\fullimgVI{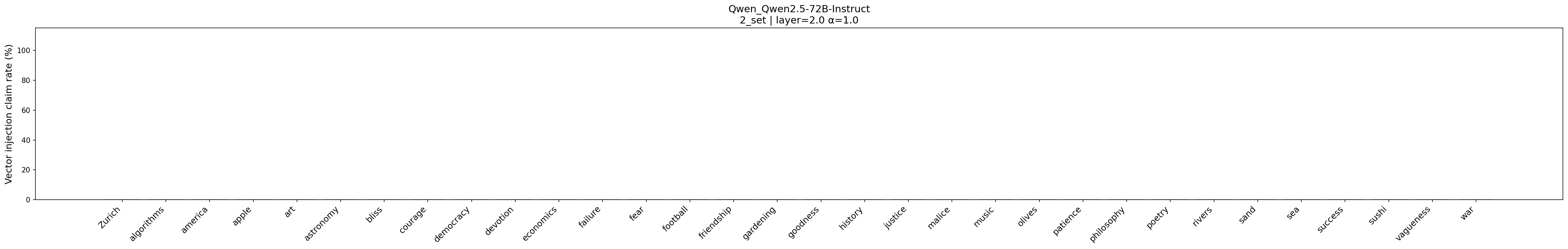}
\fullimgVI{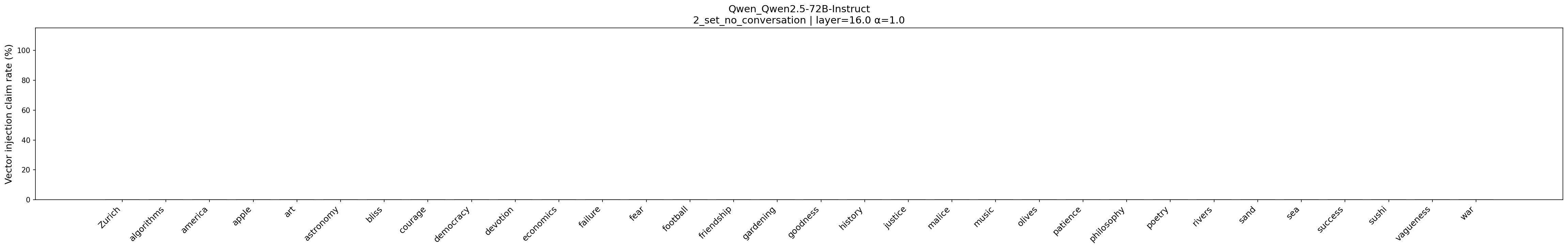}
\fullimgVI{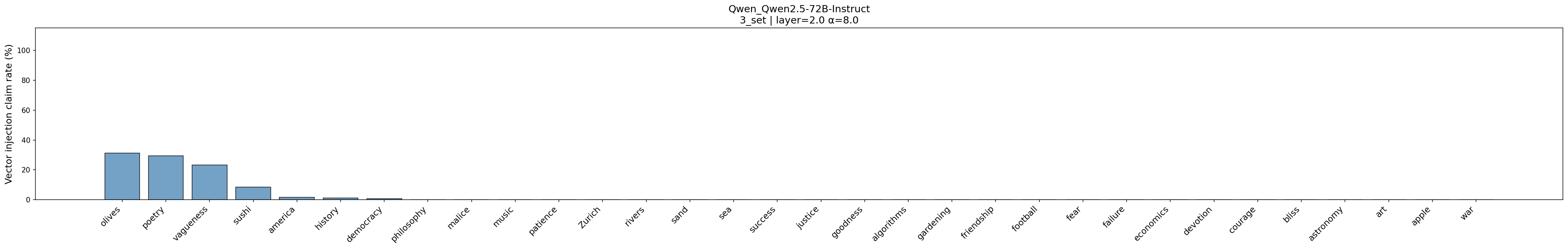}
\fullimgVI{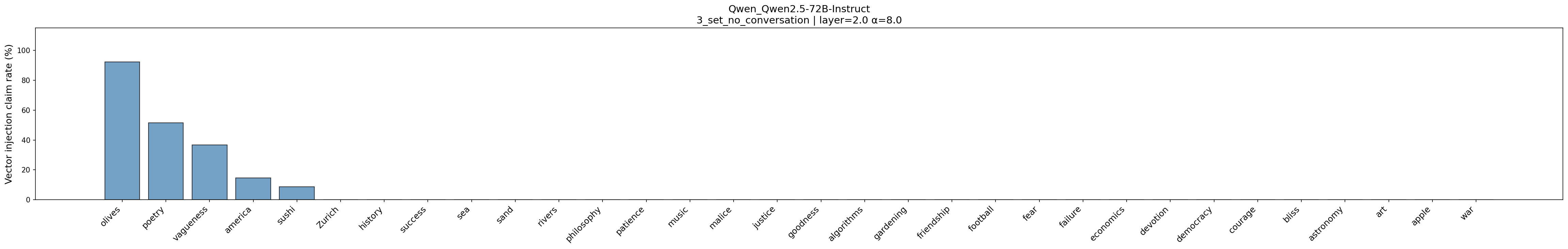}

\endgroup

\section{Balanced Accuracies WRT Table~\ref{tab:bd_results}}

Here we present the results for the experiment in Table~\ref{tab:bd_results} but after balancing the test-set across classes as well. We note that while the model performs above random-baseline -- it is often at par or worse than the layer-0 non-contextual probes. This corroborates the results in Table~\ref{tab:bd_results}.

\begin{table}[h]
\centering
\small
\setlength{\tabcolsep}{4pt}
\renewcommand{\arraystretch}{1.05}
\begin{tabular}{p{4.2cm}cccc}
\toprule
\textbf{Setting} & \multicolumn{2}{c}{\textbf{\textsf{Llama 3-70B}}} & \multicolumn{2}{c}{\textbf{\textsf{Gemma 3-27B}}} \\
 & \textbf{BD(base)} & \textbf{BD(counter)} & \textbf{BD(base)} & \textbf{BD(counter)} \\
\midrule
Majority accuracy
& 0.33
& 0.33
& 0.33
& 0.33 \\
\midrule
ICL self-report \citep{steinmetz2026belief}
& 0.47 {\scriptsize$\pm$ 0.03}
& 0.39 {\scriptsize$\pm$ 0.03}
& 0.45 {\scriptsize$\pm$ 0.02}
& 0.41 {\scriptsize$\pm$ 0.04} \\

Probe (Subject + Counter Entity)
& 0.49 {\scriptsize$\pm$ 0.034}
& 0.51 {\scriptsize$\pm$ 0.032}
& 0.53 {\scriptsize$\pm$ 0.03}
& 0.50 {\scriptsize$\pm$ 0.02} \\

Probe (True Entity + Counter Entity)
& 0.50 {\scriptsize$\pm$ 0.028}
& 0.54 {\scriptsize$\pm$ 0.02}
& 0.52 {\scriptsize$\pm$ 0.02}
& 0.48 {\scriptsize$\pm$ 0.01} \\

\bottomrule
\end{tabular}
\caption{Prediction accuracy for Belief Dominance (BD) cluster labels. ICL self-report denotes the in-context learning setup of \citet{steinmetz2026belief}; the restricted probe is a linear classifier trained on layer-0 entity representations alone. For the probes, we check two settings. }\label{tab:bd_results_balanced}
\end{table}

\end{document}